\documentclass{style}

\usepackage[utf8]{inputenc} %
\usepackage{microtype} 		%
\usepackage{enumitem} 		%
\usepackage[hyphens]{url}
\usepackage{hyphenat}		%
\usepackage{seqsplit}		%
\usepackage[english]{babel} %
\usepackage[
autostyle,
english = american
]{csquotes}					%
\MakeOuterQuote{"}

\usepackage{amsmath} 		%
\usepackage{amssymb} 		%

\usepackage{booktabs} 		%
\usepackage{longtable}		%
\usepackage{multirow}		%
\usepackage{capt-of}		%
\usepackage{tikz}

\usepackage{subfiles}		%
\usepackage[
disable
]{todonotes}
\usepackage[single=true]{acro}

\usepackage{siunitx}
\DeclareSIUnit{\pp}{\textup{pp}}

\usepackage{etoolbox}
\robustify\bfseries

\usepackage{xspace} %
\newcommand{\corpusexample}[1]{%
	\textit{"#1"}\xspace
}

\newcommand{\plhdr}{\tikz[baseline=-2pt]\fill[gray] (0,0) circle (.07em);}

\usepackage{xparse}
\NewDocumentCommand{\rot}{O{45} O{1em} m}{\makebox[#2][l]{\rotatebox{#1}{#3}}}%

\graphicspath{{img/}}

\usepackage{hyperref}
\definecolor{darkblue}{rgb}{0, 0, 0.5}
\hypersetup{colorlinks=true,citecolor=darkblue, linkcolor=darkblue, urlcolor=darkblue}

\usepackage[noabbrev]{cleveref}		%

\makeatletter
\renewcommand{\appendixsection}[1]{\refstepcounter{section}%
	\setcounter{table}{0}
	\setcounter{figure}{0}
	\setcounter{equation}{0}
	\section*{Appendix \Alph{section}: #1}%
	\def\cref@currentlabel{[appendix][\arabic{section}][]\Alph{section}}%
	\def\@currentlabelname{#1}%
}
\makeatother

\renewcommand{\appendix}{%
	\setcounter{section}{0}
	\renewcommand{\theequation}{\Alph{section}.\arabic{equation}}
	\renewcommand{\thefigure}{\Alph{section}.\arabic{figure}}
	\renewcommand{\thetable}{\Alph{section}.\arabic{table}}
	\renewcommand{\theHequation}{\Alph{section}.\arabic{equation}}
	\renewcommand{\theHfigure}{\Alph{section}.\arabic{figure}}
	\renewcommand{\theHtable}{\Alph{section}.\arabic{table}}
	\renewcommand{\theHsection}{\Alph{section}.\arabic{table}}
	\renewcommand{\theHsection}{\Alph{section}.\arabic{section}}
}

\DeclareAcronym{cdevcr}{
	short = CDCR,
	long = {cross-document event coreference resolution},
	tag = main
}
\DeclareAcronym{srl}{
	short = SRL,
	long = {semantic role labeling},
	tag = main
}
\DeclareAcronym{qa}{
	short = QA,
	long = {question answering},
	tag = main
}
\DeclareAcronym{mds}{
	short = MDS,
	long = {multi-document summarization},
	tag = main
}
\DeclareAcronym{iaa}{
	short = IAA,
	long = {inter-annotator agreement},
	tag = main
}
\DeclareAcronym{mlp}{
	short = MLP,
	long = {multi-layer perceptron},
	tag = main
}
\DeclareAcronym{wsd}{
	short = WSD,
	long = {word sense disambiguation},
	tag = main
}

\DeclareAcronym{cdevc}{
	short = CDC,
	long = {cross-document event coreference},
	tag = writing-aid
}

\DeclareAcronym{ecb}{
	short = ECB,
	long = {EventCorefBank},
	extra = {The first corpus iteration developed by \citet{bejan2010unsupervised}.},
	tag = corpus
}
\DeclareAcronym{eecb}{
	short = EECB,
	long = {Extended EventCorefBank},
	extra = {\Citet{lee2012joint}'s extension of the ECB corpus in which entity coreference annotations were added.},
	tag = corpus
}
\DeclareAcronym{ecbp}{
	short = ECB+,
	long = {EventCorefBank+},
	extra = {Extension of the EECB corpus in which \citet{cybulska2014using} added a second subtopic for each topic.},
	tag = corpus
}
\DeclareAcronym{gvc}{
	short = GVC,
	long = {Gun Violence Corpus},
	extra = {Developed by \citet{vossen2018dont}.},
	tag = corpus
}
\DeclareAcronym{fcc}{
	short = FCC,
	long = {Football Coreference Corpus},
	extra = {Sentence-level corpus developed by \citet{bugert2020breaking}.},
	tag = corpus
}
\DeclareAcronym{fcct}{
	short = FCC-T,
	long = {Football Coreference Corpus},
	extra = {Token-level reannotation of \ac{fcc} produced in this work.},
	tag = corpus
}

\DeclareAcronym{cr2020}{
	short = CR2020,
	long = \citet{cremisini2020new},
	tag = system
}
\DeclareAcronym{me2020}{
	short = ME2020,
	long = \citet{meged2020paraphrasing},
	tag = system
}
\DeclareAcronym{ba2019}{
	short = BA2019,
	long = \citet{barhom2019revisiting},
	tag = system
}
\DeclareAcronym{ke2018}{
	short = KE2018,
	long = \citet{kenyon-dean2018resolving},
	tag = system
}
\DeclareAcronym{mi2018}{
	short = MI2018,
	long = \citet{mirza2018koi},
	tag = system
}
\DeclareAcronym{vo2018}{
	short = VO2018,
	long = \citet{vossen2018newsreader},
	tag = system
}
\DeclareAcronym{vo2016}{
	short = VO2016,
	long = \citet{vossen2016identity},
	tag = system
}
\DeclareAcronym{cy2015}{
	short = CY2015,
	long = \citet{cybulska2015bag},
	tag = system
}
\DeclareAcronym{ya2015}{
	short = YA2015,
	long = \citet{yang2015hierarchical},
	tag = system
}
\DeclareAcronym{le2012}{
	short = LE2012,
	long = \citet{lee2012joint},
	tag = system
}

\date{\today}

\title{Generalizing Cross-Document Event Coreference Resolution Across Multiple Corpora}

\author{Michael Bugert}
\affil{UKP Lab\\Department of Computer Science\\Technical University of Darmstadt\\https://www.ukp.tu-darmstadt.de/}
\author{Nils Reimers}
\affil{UKP Lab}
\author{Iryna Gurevych}
\affil{UKP Lab}

\newcommand{\githuburl}{\url{https://github.com/UKPLab/cdcr-beyond-corpus-tailored}}
\newcommand{\corpusurl}{\url{https://tudatalib.ulb.tu-darmstadt.de/handle/tudatalib/2305}}

\begin{document}
\maketitle

\begin{abstract}
Cross-document event coreference resolution (CDCR) is an NLP task in which mentions of events need to be identified and clustered throughout a collection of documents.
CDCR aims to benefit downstream multi-document applications, but despite recent progress on corpora and system development, downstream improvements from applying CDCR have not been shown yet.
We make the observation that every CDCR system to date was developed, trained, and tested only on a single respective corpus. This raises strong concerns on their generalizability --- a must-have for downstream applications where the magnitude of domains or event mentions is likely to exceed those found in a curated corpus.
To investigate this assumption, we define a uniform evaluation setup involving three CDCR corpora: ECB+, the Gun Violence Corpus and the Football Coreference Corpus (which we reannotate on token level to make our analysis possible). We compare a corpus-independent, feature-based system against a recent neural system developed for ECB+. Whilst being inferior in absolute numbers, the feature-based system shows more consistent performance across all corpora whereas the neural system is hit-and-miss.
Via model introspection, we find that the importance of event actions, event time, etc. for resolving coreference in practice varies greatly between the corpora.
Additional analysis shows that several systems overfit on the structure of the ECB+ corpus.
We conclude with recommendations on how to achieve generally applicable CDCR systems in the future --- the most important being that evaluation on multiple CDCR corpora is strongly necessary.
To facilitate future research, we release our dataset, annotation guidelines, and system implementation to the public.\footnote{\githuburl}
\end{abstract}

\section{Introduction}

\begin{figure}
	\centering
	\includegraphics{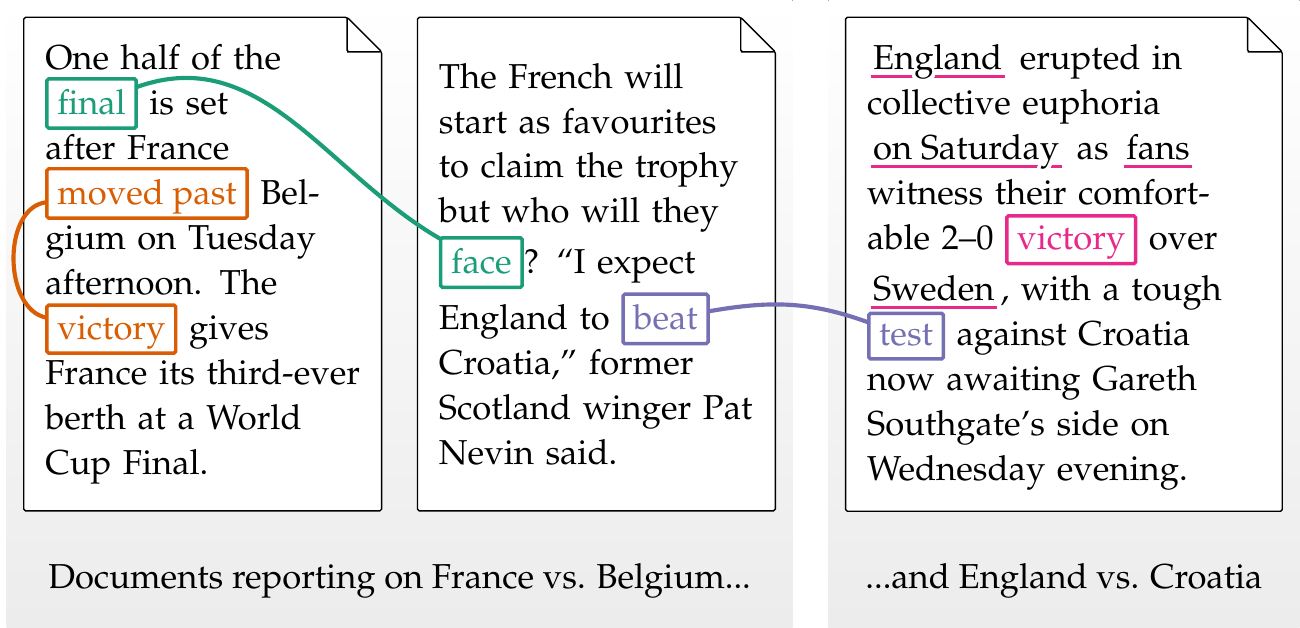}
	\caption{\Acf{cdevcr} example with excerpts of three documents from our token-level reannotation of the \acf{fcct}. The seven indicated event mentions refer to four different events. For the \corpusexample{victory} event mention, three participant mentions and one temporal mention are additionally marked.}
	\label{fig:cdevcr_example}
\end{figure}

To move beyond interpreting documents in isolation in multi-document NLP tasks such as \acl{mds} or \acl{qa}, a text understanding technique is needed to connect statements from different documents. A strong contender for this purpose is \ac{cdevcr}. In this task, systems needs to (1) find mentions of events in a collection of documents and (2) cluster those mentions together which refer to the same event (see \Cref{fig:cdevcr_example}).
An event refers to an action taking place at a certain time and location with certain participants~\cite{cybulska2014using}.
\Ac{cdevcr} requires deep text understanding and depends on a multitude of other NLP tasks such as \ac{srl}, temporal inference, and spatial inference, each of which is still being researched and not yet solved.
Furthermore, \ac{cdevcr} systems need to correctly predict the coreference relation between any pair of event mentions in a corpus. Since the number of pairs grows quadratically with the number of mentions, achieving \textit{scalable} text understanding becomes an added challenge in \ac{cdevcr}.

In recent years, new \ac{cdevcr} corpora such as the \ac{gvc}~\cite{vossen2018dont} and \ac{fcc}~\cite{bugert2020breaking} have been developed, and the state-of-the-art performance on the most commonly used corpus \acs{ecbp}~\cite{cybulska2014using} has risen steadily~\cite{meged2020paraphrasing, barhom2019revisiting, kenyon-dean2018resolving}.
We believe that \ac{cdevcr} can play a vital role for downstream multi-document tasks, and so do other authors in this area~\cite{bejan2014unsupervised, yang2015hierarchical, upadhyay2016revisiting, choubey2017event, choubey2018identifying, choubey2018improving, kenyon-dean2018resolving, barhom2019revisiting}. Yet, despite the progress made so far, we are not aware of a study which demonstrates that employing a recent \ac{cdevcr} system is indeed helpful downstream.
We make the key observation that all existing \ac{cdevcr} systems~\cite{meged2020paraphrasing, cremisini2020new, barhom2019revisiting, kenyon-dean2018resolving, mirza2018koi, vossen2018newsreader} were designed, trained, and evaluated on a single corpus respectively.
This points to a risk of systems overspecializing on their target corpus instead of learning to solve the overall task, rendering such systems unsuitable for downstream applications where generality and robustness is required. The fact that \ac{cdevcr} annotation efforts annotated only a subset of all coreference links to save costs~\cite{bugert2020breaking} further aggravates this situation.

We are, to the best of our knowledge, the first to investigate this risk.
In this work, we determine the state of generalizability in \ac{cdevcr} with respect to corpora and systems, identify the current issues, and formulate recommendations on how \ac{cdevcr} systems which are robustly applicable in downstream scenarios can be achieved in the future.
We divide our analysis into five successive stages:
\begin{enumerate}
	\item Cross-dataset modeling of \ac{cdevcr} is made difficult by annotation differences between the \ac{ecbp}, \ac{fcc}, and \ac{gvc} corpora. We establish compatibility by annotating the \acs{fcct}, an extension of the \ac{fcc} reannotated on the token level. %
	\item Analyzing generalizability across corpora is best performed with an interpretable \ac{cdevcr} system which is equally applicable on all corpora. To fulfill this requirement, we develop a conceptually simple mention-pair \ac{cdevcr} system which uses the union of features found in related work.
	\item To compare the generalization capabilities of \ac{cdevcr} \textit{system architectures}, we train and test this system and a close to state-of-the-art neural system~\cite{barhom2019revisiting} on the \ac{ecbp}, \ac{fcct} and \ac{gvc} corpora. We find that the neural system does not robustly handle \ac{cdevcr} on all corpora because its input features and architecture require \ac{ecbp}-like corpora.
	\item There is a lack of knowledge on how the \ac{cdevcr} task manifests itself in each corpus, especially with regard to which pieces of information (out of event action, participants, time, and location) are the strongest signals for event coreference. Via model introspection, we observe significant differences between corpora, finding that decisions in \ac{ecbp} are strongly driven by event actions whereas \ac{fcct} and \ac{gvc} are more balanced and additionally require text understanding of event participants and time.
	\item Finally, we evaluate our feature-based system in a cross-dataset transfer scenario to analyze the generalization capabilities of trained \ac{cdevcr} \textit{models}. We find that models trained on a single corpus do not perform well on other unseen corpora.
\end{enumerate}
Based on these findings, we conclude with recommendations for the evaluation of \ac{cdevcr} which will pave the way for more general and comparable systems in the future. Most importantly, the results of our analysis unmistakably show that evaluation on multiple corpora is imperative given the current set of available \ac{cdevcr} corpora.

\paragraph{Article Structure}
The next section provides background information on the \ac{cdevcr} task, corpora, and systems, followed by related work on feature importance in \ac{cdevcr} (\Cref{sec:related_work}). \Cref{sec:annotation} covers the re-annotation and extension of the \ac{fcc} corpus. %
We explain the feature-based \ac{cdevcr} system in \Cref{sec:system} before moving on to a series of experiments: we compare this system and the neural system of \citet{barhom2019revisiting} in \Cref{sec:exp_in_dataset}. In \Cref{sec:exp_signals} we analyze the signals for event coreference in each corpus. Lastly, we test model generalizability across corpora in \Cref{sec:exp_cross_dataset}. We discuss the impact of these experiments and offer summarized recommendations on how to achieve general \ac{cdevcr} systems in the future in \Cref{sec:discussion,sec:future_work}. We conclude with \Cref{sec:conclusion}.

\section{Background on \ac{cdevcr}}
\label{sec:background}

We explain the \ac{cdevcr} task in greater detail, report on the most influential \ac{cdevcr} datasets and cover notable coreference resolution systems developed for each corpus.

\subsection{Task Definition}
\label{sec:cdevcr_task_definition}

The \ac{cdevcr} task is studied for several domains including news events in (online) news articles, events pertaining to the treatment of patients in physician's notes~\cite{raghavan2014cross,wright-bettner2019cross} or the identification and grouping of biomedical events in research literature~\cite{vanlandeghem2013large}.
In this work, we restrict ourselves to the most explored variant of \acl{cdevcr} in the news domain.

We follow the task definition and terminology of \citet{cybulska2014using}. Here, events consist of four event components -- an action, several human or non-human participants, a time and a location. Each of these components can be mentioned in text, that is, an \textit{action mention} would be the text span referencing the action of an event instance. An example is shown in \Cref{fig:cdevcr_example}, where the rightmost document references a football match between England and Sweden. The action mention for this event is \corpusexample{victory}, alongside three entity mentions \corpusexample{England} (the population of England), \corpusexample{fans} (English football fans), and \corpusexample{Sweden} (the Swedish national football team) who took part in the event. The temporal expression \corpusexample{on Saturday} grounds the event mention to a certain time, which in this case depends on the date the news article was published on.

Different definitions have been proposed for the relation of event coreference.
Efforts such as ACE~\cite{walker2006ace} only permit the annotation of identity between event mentions whereas \citet{hovy2013events} further distinguish subevent or membership relations.
Definitions generally need to find a compromise between complexity and ease of annotation, particularly for the cross-document case (see \citet{wright-bettner2019cross} for a detailed discussion).
We follow the (comparatively simple) definition of \citet{cybulska2014using} in which two action mentions corefer if they refer to the same real-world event, meaning their actions and their associated participants, time, and location are semantically equivalent. Relevant examples are shown in \Cref{fig:cdevcr_example} where all action mentions of the same color refer to the same event.
The two steps a \ac{cdevcr} system needs to perform therefore are (1) the detection of event actions and event components and (2) the disambiguation of event actions to produce a cross-document event clustering.
A challenging aspect of \ac{cdevcr} is the fact that finding mentions of all four event components in the same sentence is rare, meaning that information may have to be inferred from the document context or in some cases, it may not be present in the document at all. The second challenge is efficiently scaling the clustering process to large document collections with thousands of event mentions since every possible pair of event mentions could together form a valid cluster.

\subsection{System Requirements}
\label{sec:cdevcr_system_requirements}
The requirements that downstream applications place on systems resolving \acl{cdevc} can be diverse. We establish high-level requirements that a system performing \ac{cdevcr} on news text should meet:
\begin{itemize}
	\item Datasets may consist of many interwoven topics. Systems should perform well on a \textbf{broad selection of event types} of different properties (punctual events such as accidents, longer-term events such as natural disasters, pre-planned events such as galas or sports competitions).
	\item To provide high-quality results, systems should \textbf{fully support the definition of event coreference} mentioned previously, meaning they find associations between event mentions at a level human readers would be able to by inferring temporal and spatial clues from the document context and reasoning over event action and participants.
	\item Datasets may consist of a large number of documents containing many event mentions. We expect \ac{cdevcr} systems to be \textbf{scalable} enough to handle 100k event mentions in a reasonable amount of time (less than one day on a single-GPU workstation).
\end{itemize}

\subsection{Corpora}
\label{sec:related_work_datasets}

The corpus most commonly associated with \ac{cdevcr} is \textbf{\acf{ecbp}}. Originally developed as the \ac{ecb} corpus~\cite{bejan2010unsupervised}, it was enriched with entity coreference annotations by \citet{lee2012joint} to form the \ac{eecb} corpus. This corpus was later extended with 500 additional documents by \citet{cybulska2014using} to create the \ac{ecbp} corpus.
This most recent version contains 982 news articles on 43 topics. The topics were annotated separately, meaning there are no coreference links across topics. For each topic (for example \corpusexample{bank explosions}), there are two main events (\corpusexample{Bank explosion in Oregon 2008} and \corpusexample{Bank explosion in Athens 2012}) and several news documents which report on either of those two events.
The set of documents reporting on the same event is commonly referred to as a \textit{subtopic}.
\ac{ecbp} is the only corpus of those discussed here which does not provide the publication date for each document. It does however contain annotations for all four event components as well as additional cross-document entity coreference annotations for participants, time, and location mentions.

\acuse{eecb}

The \textbf{\acf{fcc}}~\cite{bugert2020breaking} contains 451 sports news articles on football tournaments annotated with \acl{cdevc}. The annotation was carried out via crowdsourcing and focused on retrieving \textit{cross-subtopic} event coreference links.
Following the nomenclature of \citet{bugert2020breaking}, a within-subtopic coreference link is defined by a pair of coreferring event mentions which originate from two documents reporting about the same overall event. For example in \ac{ecbp}, two different news articles reporting about the same bank explosion in Athens in the year 2012 may both mention the event of the perpetrators fleeing the scene. For a \textit{cross-subtopic} event coreference link, two event mentions from articles on \textit{different} events need to corefer. A sports news article summarizing a quarter-final match of a tournament could for example recommend watching the upcoming semifinal, whereas an article written weeks later about the grand final may refer to the same semifinal in an enumeration of a team's past performances in the tournament. A concrete example is shown in \Cref{fig:cdevcr_example}, where the mentions \corpusexample{beat} and \corpusexample{test} corefer while belonging to different subtopics. Cross-subtopic coreference links are a crucial aspect of \ac{cdevcr} since they connect mentions from documents with low content overlap, forming far-reaching coreference clusters which should prove particularly beneficial for downstream applications~\cite{bugert2020breaking}.
In \ac{fcc}, event mentions are annotated only at the sentence level contrary to \ac{ecbp} and \ac{gvc} which feature token level annotations.

The \textbf{\acf{gvc}}~\cite{vossen2018dont} is a collection of 510 news articles covering 241 gun violence incidents. The goal was to create a challenging \ac{cdevcr} corpus with many similar event mentions. Each news article belongs to the same topic (gun violence) and only event mentions related to gun violence were annotated (\corpusexample{kill}, \corpusexample{wounded}, etc.). Cross-subtopic coreference links were not annotated.

\begin{table}
	\centering
	
	\newsavebox\checkmarkBox
	\sbox{\checkmarkBox}{\checkmark}
	\newsavebox\checkmarkParensBox
	\sbox{\checkmarkParensBox}{(\checkmark)}
	\newsavebox{\plhdrBox}
	\savebox{\plhdrBox}[\wd\checkmarkBox][c]{\plhdr}
	\newsavebox{\checkmarkBoxParensSpaced}
	\savebox{\checkmarkBoxParensSpaced}[\wd\checkmarkParensBox][c]{\checkmark}
	\newsavebox{\plhdrBoxParensSpaced}
	\savebox{\plhdrBoxParensSpaced}[\wd\checkmarkParensBox][c]{\plhdr}
	
	\begin{tabular}{@{}l*{2}{S[table-format=5.0, table-text-alignment=right]}S[table-format=7.0, table-text-alignment=right]S[table-format=6.0, table-text-alignment=right]@{}}
		\toprule
		& {\acs{ecbp}}             & {\acs{gvc}}              & {\acs{fcc}}              & {\acs{fcct}}                         \\
		\midrule
		annotation unit                          & {token}                  & {token}                  & {sentence}               & {token}                              \\
		included annotations                     &                          &                          &                          &                                      \\
		\hspace{.75em} event coreference         & {\usebox{\checkmarkBox}} & {\usebox{\checkmarkBox}} & {\usebox{\checkmarkBox}} & {\usebox{\checkmarkBoxParensSpaced}} \\
		\hspace{.75em} entity coreference        & {\usebox{\checkmarkBox}} & {\usebox{\plhdrBox}}     & {\usebox{\plhdrBox}}     & {\usebox{\plhdrBoxParensSpaced}}     \\
		\hspace{.75em} document publ. date       & {\usebox{\plhdrBox}}     & {\usebox{\checkmarkBox}} & {\usebox{\checkmarkBox}} & {\usebox{\checkmarkBoxParensSpaced}} \\
		\hspace{.75em} particip., time, location & {\usebox{\checkmarkBox}} & {\usebox{\plhdrBox}}     & {\usebox{\plhdrBox}}     & {\usebox{\checkmarkBoxParensSpaced}} \\
		\hspace{.75em} semantic roles            & {\usebox{\plhdrBox}}     & {\usebox{\plhdrBox}}     & {\usebox{\plhdrBox}}     & {\usebox{\checkmarkParensBox}}       \\
		\addlinespace
		topics                                   & 43                       & 1                        & 1                        & 1                                    \\
		subtopics per topic                      & 2                        & 241                      & 183                      & 183                                  \\
		documents                                & 982                      & 510                      & 451                      & 451                                  \\
		sentences                                & 16314                    & 9782                     & 14940                    & 14940                                \\
		event mentions                           & 6833                     & 7298                     & 2374                     & 3563                                 \\
		event clusters                           & 2741                     & 1411                     & 218                      & 469                                  \\
		\hspace{.75em} singletons                & 2019                     & 365                      & 50                       & 185                                  \\
		entities / event comp.                   &                          &                          &                          &                                      \\
		\hspace{.75em} participant               & 12676                    & {n/a}                    & {n/a}                    & 5937                                 \\
		\hspace{.75em} time                      & 2412                     & {n/a}                    & {n/a}                    & 1439                                 \\
		\hspace{.75em} location                  & 2205                     & {n/a}                    & {n/a}                    & 566                                  \\
		event coreference links                  & 26712                    & 29398                    & 106479                   & 145272                               \\
		\hspace{.75em} within-document           & 1636                     & 14218                    & 2344                     & 2662                                 \\
		\hspace{.75em} within-subtopic           & 24816                    & 15180                    & 3972                     & 4561                                 \\
		\hspace{.75em} cross-subtopic            & 260                      & 0                        & 100163                   & 138049                               \\
		\hspace{.75em} cross-topic               & 0                        & 0                        & 0                        & 0                                    \\
		\bottomrule
	\end{tabular}
	\caption{Comparison of annotations in several \ac{cdevcr} corpora. Values for \acs{fcc} refer to our cleaned version of the original corpus. \acs{fcct} is our token-level reannotation.}
	\label{tab:dataset_properties}
\end{table}

\begin{figure}
	\centering
	\includegraphics{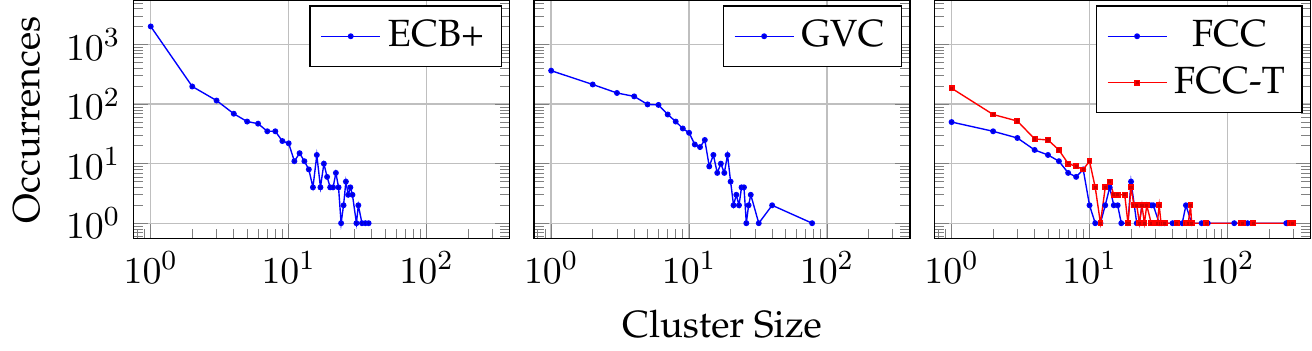}
	\caption{Cluster size distribution in \ac{cdevcr} corpora.}
	\label{fig:cluster_size_distribution}
\end{figure}

\Cref{tab:dataset_properties} presents further insights into these corpora. There, we report the total number of event coreference links in each corpus and categorize them by type. Note that in \ac{ecbp} and \ac{gvc}, nearly all cross-document links are of the within-subtopic kind whereas \ac{fcc} focused on annotating cross-subtopic links. The stark contrast in the number of coreference links between \ac{fcc} and \ac{ecbp}/\ac{gvc} can be attributed to the facts that (1) the number of coreference links grows quadratically with the number of mentions in a cluster and (2) \ac{fcc} contains clusters with more than 100 mentions, see~\Cref{fig:cluster_size_distribution}.

While the annotation design of each of these corpora has had different foci, they share commonalities. The structure of each corpus can be framed as a hierarchy with three levels: there are one or more \textit{topics/event types} which each contain \textit{subtopics/event instances} which each contain multiple \textit{documents}. Both \ac{ecbp} and \ac{gvc} annotate event mentions on the token level in a similar manner.
Since \ac{fcc} is the only \ac{cdevcr} corpus missing token level event mention annotations, we add these annotations in this work to produce the \ac{fcct} corpus (see \Cref{sec:annotation}).
With this change made, it is technically and theoretically possible to examine these \ac{cdevcr} corpora jointly.

\subsection{Systems}
We here summarize the principles of \ac{cdevcr} systems, followed by the state-of-the-art systems for each \ac{cdevcr} corpus.

\subsubsection{System Principles}
Given a collection of event mentions, a discrete or vectorized representation needs to be created for each mention so that the mentions can be clustered. Following the definition of the \ac{cdevcr} task, a representation should contain information on the action, participants, time, and location of the event mention. This information may be scattered throughout the document and needs to be extracted first.
To do this, \ac{cdevcr} may preprocess documents via \ac{srl}, temporal tagging or entity linking.

Two general strategies exist for computing the distances between mentions which are needed for clustering: representation learning and metric learning~\cite{hermans2017defense}. \textbf{Representation learning} approaches produce a vector representation for each event mention independently. The final event clustering is obtained by computing the cosine distance between each vector pair, followed by agglomerative clustering on the resulting distance matrix.
Most approaches belong to the group of conceptually simpler \textbf{metric learners} which predict the semantic distance between two mentions or clusters based on a set of features. By applying the metric on all $\binom{n}{2}$ pairs for $n$ mentions, a distance matrix is obtained which is then fed to a clustering algorithm. Any probabilistic classifier or regression model may be used to obtain the mention distances.
Metric learning approaches can be further divided into \textbf{mention pair} approaches which compute the distance between each mention pair once and \textbf{cluster pair} approaches which recompute cluster representations and distances after each cluster merge.
Computing the distance between all mention pairs can be a computationally expensive process. Some metric learner approaches therefore perform a separate document preclustering step to break down the task into manageable parts. The metric learning approach is then applied on each individual cluster of documents and its results are combined to produce the final coreference clustering.

\begin{table}
	\small
	\centering
	\setlength{\tabcolsep}{2.5pt}
	\newcommand{\tang}{45}		%
	\newcommand{\twid}{1.7em}	%
	\begin{tabular}{@{}ll*{11}{c}@{}}
		\toprule
		&                             & \rot[\tang][\twid]{\acs{cr2020}} & \rot[\tang][\twid]{\acs{me2020}} & \rot[\tang][\twid]{\acs{ba2019}} & \rot[\tang][\twid]{\acs{ke2018}} & \rot[\tang][\twid]{\acs{vo2016}} & \rot[\tang][\twid]{\acs{cy2015}} & \rot[\tang][\twid]{\acs{ya2015}} & \rot[\tang][\twid]{\acs{le2012}} & \rot[\tang][\twid]{\acs{mi2018}} & \rot[\tang][\twid]{\acs{vo2018}} & \rot[\tang][\twid]{\textbf{ours}} \\
		\midrule
		\textbf{Prepro-}    & Fact KB entity linking      &              \plhdr              &              \plhdr              &              \plhdr              &              \plhdr              &            \checkmark            &              \plhdr              &              \plhdr              &              \plhdr              &            \checkmark            &            \checkmark            &            \checkmark             \\
		\textbf{cessing}    & Lexical KB entity linking   &              \plhdr              &              \plhdr              &              \plhdr              &              \plhdr              &            \checkmark            &            \checkmark            &            \checkmark            &            \checkmark            &            \checkmark            &            \checkmark            &              \plhdr               \\
		& \Acl{srl}                   &              \plhdr              &            \checkmark            &            \checkmark            &              \plhdr              &            \checkmark            &              \plhdr              &            \checkmark            &            \checkmark            &              \plhdr              &            \checkmark            &            \checkmark             \\
		& Temporal tagging            &              \plhdr              &              \plhdr              &              \plhdr              &              \plhdr              &            \checkmark            &              \plhdr              &              \plhdr              &              \plhdr              &              \plhdr              &            \checkmark            &            \checkmark             \\
		\midrule
		\textbf{Mention or} & Bag of words                &              \plhdr              &              \plhdr              &              \plhdr              &            \checkmark            &              \plhdr              &              \plhdr              &              \plhdr              &            \checkmark            &              \plhdr              &              \plhdr              &              \plhdr               \\
		\textbf{Document}   & TF--IDF                     &            \checkmark            &              \plhdr              &              \plhdr              &            \checkmark            &              \plhdr              &              \plhdr              &            \checkmark            &              \plhdr              &              \plhdr              &            \checkmark            &            \checkmark             \\
		\textbf{Represent-} & Word emb. simple            &            \checkmark            &            \checkmark            &            \checkmark            &            \checkmark            &              \plhdr              &              \plhdr              &            \checkmark            &              \plhdr              &              \plhdr              &              \plhdr              &              \plhdr               \\
		\textbf{ations}     & Word emb. contextual        &              \plhdr              &            \checkmark            &            \checkmark            &              \plhdr              &              \plhdr              &              \plhdr              &              \plhdr              &              \plhdr              &              \plhdr              &              \plhdr              &            \checkmark             \\
		& Character embeddings        &              \plhdr              &            \checkmark            &            \checkmark            &              \plhdr              &              \plhdr              &              \plhdr              &              \plhdr              &              \plhdr              &              \plhdr              &              \plhdr              &              \plhdr               \\
		\midrule
		\textbf{Features}   & Entity Coreference          &              \plhdr              &            \checkmark            &            \checkmark            &              \plhdr              &              \plhdr              &            \checkmark            &              \plhdr              &            \checkmark            &              \plhdr              &              \plhdr              &              \plhdr               \\
		& Ling. properties of mention &            \checkmark            &           (\checkmark)           &           (\checkmark)           &              \plhdr              &              \plhdr              &              \plhdr              &              \plhdr              &            \checkmark            &              \plhdr              &              \plhdr              &              \plhdr               \\
		& Paraphrase detection        &              \plhdr              &            \checkmark            &              \plhdr              &              \plhdr              &              \plhdr              &              \plhdr              &              \plhdr              &              \plhdr              &              \plhdr              &              \plhdr              &              \plhdr               \\
		& Temporal distance           &              \plhdr              &              \plhdr              &              \plhdr              &              \plhdr              &            \checkmark            &              \plhdr              &              \plhdr              &              \plhdr              &            \checkmark            &            \checkmark            &            \checkmark             \\
		& Spatial distance            &              \plhdr              &              \plhdr              &              \plhdr              &              \plhdr              &              \plhdr              &              \plhdr              &              \plhdr              &              \plhdr              &              \plhdr              &              \plhdr              &            \checkmark             \\
		& Compare discrete reprs.     &              \plhdr              &              \plhdr              &              \plhdr              &              \plhdr              &            \checkmark            &            \checkmark            &            \checkmark            &            \checkmark            &            \checkmark            &              \plhdr              &            \checkmark             \\
		& Compare vectorized reprs.   &           (\checkmark)           &           (\checkmark)           &           (\checkmark)           &            \checkmark            &              \plhdr              &              \plhdr              &            \checkmark            &            \checkmark            &              \plhdr              &            \checkmark            &            \checkmark             \\
		& Compare w.r.t. lexical KB   &              \plhdr              &              \plhdr              &              \plhdr              &              \plhdr              &            \checkmark            &            \checkmark            &              \plhdr              &            \checkmark            &            \checkmark            &              \plhdr              &              \plhdr               \\
		& Discourse-related features  &              \plhdr              &              \plhdr              &              \plhdr              &            \checkmark            &              \plhdr              &            \checkmark            &              \plhdr              &              \plhdr              &              \plhdr              &              \plhdr              &              \plhdr               \\
		\bottomrule
	\end{tabular}
	\caption{Preprocessing steps, representations and features used by \ac{cdevcr} systems. We mark implictly learnt neural features with (\checkmark).}
	\label{tab:feature_comparison_related_work}
\end{table}

Common types of features used by \ac{cdevcr} systems are text similarity features (string matching between event mention actions), semantic features (the temporal distance between mentions), features using world knowledge (the spatial distance between the locations of mentions) or discourse features (the position of a mention in the document) as well as latent neural features.\footnote{See~\citet{lu2018event} for more examples of common features.} \Cref{tab:feature_comparison_related_work} shows the types of features which existing \ac{cdevcr} systems rely on.

\subsubsection{Notable \acs{cdevcr} Systems}
\label{sec:notable_cdevcr_systems}

\begin{table*}
	\small
	\centering
	\setlength{\tabcolsep}{3pt}
	\begin{tabular}{@{}*{7}{l}@{}}
		\toprule
		System			& Target		& Mention dist. 		& Approach		& Precluster 		& Learning				& Clustering \\
		& corpus		& computation			&				& documents?		& approach				& technique \\
		\midrule
		\acs{cr2020}	& \acs{ecbp}	& classifier			& mention pair	& yes				& \acs{mlp}				& transitive closure \\
		\acs{me2020} 	& \acs{ecbp}	& classifier			& cluster pair	& yes				& \acs{mlp}				& agglomerative \\
		\acs{ba2019} 	& \acs{ecbp}	& classifier			& cluster pair	& yes				& \acs{mlp}				& agglomerative \\
		\acs{ke2018}	& \acs{ecbp}	& representation		& mention pair	& no				& \acs{mlp} autoenc.	& agglomerative \\
		\acs{vo2016}	& \acs{ecbp}	& classifier			& mention pair	& no				& rule-based			& transitive closure \\
		\acs{cy2015}	& \acs{ecbp}	& classifier			& mention pair	& yes \& no			& decision tree			& transitive closure \\
		\acs{ya2015}	& \acs{ecbp}	& classifier			& mention pair	& no				& logistic regr.		& HDDCRP \\
		\acs{le2012}	& \acs{ecbp}	& classifier			& cluster pair	& yes				& linear regr.			& agglomerative \\
		\addlinespace
		\acs{mi2018}	& \acs{gvc}		& classifier			& document pair	& n/a				& rule-based			& agglomerative \\
		\acs{vo2018}	& \acs{gvc}		& classifier			& document pair	& n/a				& rule-based			& transitive closure \\
		\addlinespace
		\textbf{ours}	& n/a			& classifier			& mention pair	& no				& XGBoost				& agglomerative \\
		\bottomrule
	\end{tabular}
	\caption{Core principles of several \ac{cdevcr} systems}
	\label{tab:related_work_system_properties}
\end{table*}

\Cref{tab:related_work_system_properties} shows a comparison of the core principles of several \ac{cdevcr} systems in terms of their mention distance computation, learning approach and more. We compare the systems of \citealt{cremisini2020new} (\acs{cr2020}), \citealt{meged2020paraphrasing} (\acs{me2020}), \citealt{barhom2019revisiting} (\acs{ba2019}), \citealt{kenyon-dean2018resolving} (\acs{ke2018}), \citealt{vossen2016identity} (\acs{vo2016}), \citealt{cybulska2015bag} (\acs{cy2015}), \citealt{yang2015hierarchical} (\acs{ya2015}), \citealt{lee2012joint} (\acs{le2012}), \citealt{mirza2018koi} (\acs{mi2018}), and \citealt{vossen2018newsreader} (\acs{vo2018}).
We emphasize notable systems for each corpus.

At the time of writing, the state-of-the-art system on \textbf{\ac{ecbp}} is \ac{me2020}, a cluster-pair approach in which a \ac{mlp} is trained to jointly resolve entity and event coreference. It is an extension of \ac{ba2019}, adding paraphrasing features. The system performs document preclustering prior to the coreference resolution step.

\textbf{\ac{gvc}} was used in SemEval 2018 Task 5 which featured a \ac{cdevcr} subtask~\cite{postma2018semeval}. The best performing system was \ac{mi2018} which clusters documents using the output of a \ac{wsd} system, person and location entities and event times. Based on the assumption that each document mentions up to one event of each event type, the system puts all event mentions of same event type in the same cross-document event coreference cluster. Due to the nature of the shared task, the system is specialized on a limited number of event types. \acs{vo2016} and \acs{vo2018} are based on the NewsReader pipeline which contains several preprocessing stages to perform event mention detection, entity linking, \ac{wsd} and more. Using this information, one rule-based system was defined per corpus (\ac{ecbp} and \acs{gvc}) which is tailored to the topics and annotations present in the respective corpus.

The \textbf{\ac{fcc}} is the most recently released corpus of the three. We are not aware of any publications reporting results for this corpus.

\subsubsection{On the Application of Event Mention Detection}
\label{sec:mention_detection_argument}
With respect to the two steps a \ac{cdevcr} system needs to perform (event mention detection and event coreference resolution), several authors have decided to omit the first step and work on gold mentions alone~\cite{cybulska2015bag,kenyon-dean2018resolving,barhom2019revisiting,meged2020paraphrasing} which simplifies the task and system development.
Systems which include a mention detection step~\cite{lee2012joint,yang2015hierarchical,vossen2016identity,choubey2017event,vossen2018newsreader,cremisini2020new} are more faithful to the task but risk introducing another source of error. Compared to using gold event mentions, performance drops from \SI{20}{\pp} CoNLL F1~\cite{vossen2016identity} to \SI{40}{\pp} CoNLL F1~\cite{cremisini2020new} have been observed on \ac{ecbp}.
\Citeauthor{vossen2016identity} derive from these results that event detection "is the most important factor for improving event coreference"~\cite[p. 518]{vossen2016identity}.

We think that the root cause for these losses in performance are not the event detection approaches themselves but rather intentional limitations in the event mention annotations of \ac{cdevcr} corpora.
We take the \ac{ecbp} corpus as an example. Based on the event definition stated in the annotation guidelines, several hundred event mentions would qualify for annotation in each news document. To keep the annotation effort manageable, only event mentions of the document's seminal event (the main event the article is reporting about) and mentions of other events in the same sentence were annotated~\cite[p. 9]{cybulska2014guidelines}.
Conversely, the corpus contains a large amount of valid event mentions which were deliberately left unannotated.\footnote{In \SI{88}{\percent} of all sentences in \ac{ecbp}, no event actions are annotated.}
A mention detection system will (unaware of this fact) predict these event mentions anyway and will be penalized for producing false positive predictions. In the subsequent mention clustering step, coreference chains involving these surplus mentions increase the risk of incorrect cluster merges between valid mentions and will overall lead to lower precision.
A general purpose mention detection system may perform poorly on the \ac{fcc} and \ac{gvc} corpora in similar fashion. For these corpora, affordability of the annotation process was achieved by restricting event mentions to certain action types, which lowers the overall number of to-be-annotated event mentions.

We therefore think that, as long as no \ac{cdevcr} corpus exists in which every single event mention is annotated, event detection and event coreference resolution should be treated separately, meaning event coreference resolution performance should be reported on gold event mentions. For this reason, and because of the different approaches for limiting the number of event mentions in each of the three corpora, we perform all experiments on gold event mention spans in this work.

\section{Related Work}
\label{sec:related_work}
Prior work has examined feature importance in \ac{cdevcr} systems.
\Citet{cybulska2015bag} tested different combinations of features with a decision tree classifier on \ac{ecbp}. They find that system performance majorly stems from a lemma overlap feature and that adding discourse, entity coreference and \ac{wsd} features improves BLANC F1 by only \SI{1}{\pp}.
\citet{cremisini2020new} conducted a study in which they built a feature-based mention pair approach for \ac{ecbp} to gain deeper insights into the importance of features and on the performance impact of document preclustering. Amongst four features (fastText~\cite{bojanowski2017enriching} word embedding similarity between event actions, event action word distribution, sentence similarity and event action part-of-speech comparison), the embedding similarity feature was found to be the most important by far. The use of document preclustering caused an improvement of \SI{3}{\pp} CoNLL F1, leading \citeauthor{cremisini2020new} to encourage future researchers in this field to report experiments with and without document preclustering.

Our work significantly deepens these earlier analyses.
Since research on \ac{cdevcr} systems has so far only focused on resolving \acl{cdevc} in individual corpora, we tackle the issue of generalizability \textbf{across multiple corpora}. We use a \textbf{broader} set of features and by comparing two \ac{cdevcr} approaches, while previous work focused on the \ac{ecbp} corpus using the aforementioned smaller sets of features.
We (1) develop a general feature-based \ac{cdevcr} system, (2) apply it on each of the corpora mentioned above, and (3) analyze the information sources in each corpus which are most informative to \acl{cdevc}. We thereby provide the first comparative study of \ac{cdevcr} approaches, paving the way for general resolution of \acl{cdevc} which will aid downstream multi-document tasks.

\section{\ac{fcc} Reannotation}
\label{sec:annotation}
We reannotate the \acf{fcc} to improve its interoperability with \ac{ecbp} and the \acf{gvc}.\footnote{We additionally conducted an annotation of the missing document publication dates in \ac{ecbp}, but found that dates could only be manually extracted in half of the corpus documents. We therefore did not include these annotations in our experiments. More details on this annotation effort are reported in \Cref{sec:app_annotation_ecbp}.}

The \ac{fcc} was recently introduced by~\citet{bugert2020breaking} as a \ac{cdevcr} corpus with sentence-level event mention annotations (see \Cref{sec:related_work_datasets}). We reannotate all event mentions on token level, add annotations of event components, and annotate additional event mentions to produce the \textbf{\acs{fcct}} corpus (T for token level).
The following sections cover our annotation approach, inter-annotator agreement and the properties of the resulting corpus.

\subsection{Annotation Task Definition}
In the original \ac{fcc} annotation, crowd annotators were given a predefined set of events and sentences of news articles to work on. Each sentence had to be marked with the subset of events referenced in the sentence.
We take these sentences and annotate the action mention of each referenced event on token level. For each event, we additionally annotate the corresponding participants, time, and location mentions appearing in the same sentence as the action mention. To achieve maximum compatibility with existing corpora, we adopted the \ac{ecbp} annotation guidelines~\cite{cybulska2014guidelines}.\footnote{For details and examples, please refer to the guidelines published at \githuburl{}}
We distinguish between different subtypes of participants (person, organization, etc.), time, and location as done by \citet{cybulska2014guidelines}. We do not differentiate between action types since all events (pre\nobreakdash-)annotated in \ac{fcc} should belong to the \texttt{OCCURRENCE} type \citep[see][p.14]{cybulska2014guidelines}. We do not annotate (cross-document) entity coreference. We do annotate a rudimentary kind of semantic roles which we found are crucially missing in \ac{ecbp}: we instruct annotators to link mentions of participants, time, and location to their corresponding action mention.

While developing the guidelines, we noticed cases where sentence-level mentions are evidently easier to work with than token-level mentions.
For example, enumerations or aggregated statements over events (such as \corpusexample{Switzerland have won six of their seven meetings with Albania, drawing the other.}) are difficult to break down into token-level event mentions. Cases like these are not covered by the \ac{ecbp} annotation guidelines and were removed in the conversion process.
A similar issue is caused by coordinate structures such as \corpusexample{Germany beat Algeria and France in the knockout stages} where two football match events are referenced by the same verb. To handle these cases, we annotated two separate event mentions sharing the same action mention (\corpusexample{beat}). Since superimposed mention spans are not supported by coreference evaluation metrics, we additionally provide a version of the corpus in which these mentions are removed.

In \ac{fcc}, crowdworkers identified a further \num{1100} sentences which mention one or more football-related events outside of the closed set of events they were provided with during the annotation. These event mentions were left unidentified by~\citet{bugert2020breaking}. We instructed annotators to manually link each event mention in this extra set of sentences to a database of 40k international football matches\footnote{\url{https://www.kaggle.com/martj42/international-football-results-from-1872-to-2017/version/5}} and again marked and linked the token spans of actions, participants, times, and locations.

Annotators were given the option to mark sentences they found unclear or which were incorrectly annotated by crowdworkers in the original dataset. We manually resolved the affected sentences on a case-by-case basis.

\subsection{Annotation Procedure and Results}
\label{sec:annotation_results}

The annotation was carried out with the INCEpTION annotation tool~\cite{klie2018inception}.
We trained two student annotators on a set of 10 documents. The students were given feedback on their work and afterwards annotated a second batch of 22 documents independently.
\Cref{tab:iaa} shows the inter-annotator agreement on this second batch. We report Krippendorff's $\alpha_U$~\cite{krippendorff1995reliability} which measures the agreement in span overlap on character level as the micro average over all documents.

\begin{table}
	\centering
	\begin{tabular}{@{}lr@{}}
		\toprule
		action mentions									& 0.80 \\
		participants, time, location (spans only)		& 0.67 \\
		participants, time, location (incl. subtype)	& 0.57 \\
		\bottomrule
	\end{tabular}
	\caption{Inter-annotator agreement ($\alpha_U$)}
	\label{tab:iaa}
\end{table}

For the annotation of action mention extents, which is the most important step in our re-annotation effort, we reach 0.80 $\alpha_U$, indicating good reliability between annotators~\cite{carletta1996assessing, artstein2008inter}.
The agreement for the annotation of participants, time, and location is lower at 0.57 $\alpha_U$. We found that this mostly stems from the annotation of participants: In the guidelines, we specify that annotators should only mark an entity as a participant of an event if it plays a significant role in the event action. The larger and more coarse an event is, the more difficult this decision becomes for annotators. One such case is shown in \Cref{example:difficult_annotation} where it is debatable if \corpusexample{Christian Teinturier} is or is not significantly involved in the tournament event.
\begin{example}
	\corpusexample{Earlier today, French Football Federation vice-president Christian Teinturier said if there was any basis to the reports about Anelka then he should be sent home from the \textbf{tournament} immediately.}
	\label{example:difficult_annotation}
\end{example}
\noindent
A second reason is that we do not annotate entity coreference, so only a single entity mention is meant to be annotated for each entity participating in an event. In case the same entity appears twice in a sentence, we instruct annotators to choose the more specific description. If the candidates are identical in surface form, annotators are meant to choose the one closer (in word distance) to the event action. There remains a level of subjectivity in these decisions, leading to disagreement.

Overall, we concluded that the annotation methodology produced annotations of sufficient quality. The remaining 419 documents were divided among both annotators. The corpus re-annotation required 120 working hours from annotators (including training and the burn-in test).
We fixed a number of incorrect annotations in the crowdsourced \ac{fcc} corpus. For example, we removed several mentions of generic events (\corpusexample{winning a World Cup final is every player's dream}) which were incorrectly marked as referring to a concrete event.

\Cref{tab:dataset_properties} (on \cpageref{tab:dataset_properties}) shows the properties of the resulting \acs{fcct} corpus alongside \ac{ecbp}, \ac{gvc} and our cleaned version of the sentence-level \ac{fcc} corpus.
Compared to the original \ac{fcc} corpus, our token-level reannotation offers \SI{50}{\percent} more event mentions and twice as many annotated events. With respect to the \ac{srl} annotations, we analyzed how frequently event components of each type were attached to action mentions. We found that \SI{95.7}{\percent} of action mentions have at least one participant attached, \SI{41.6}{\percent} at least one time mention and \SI{15.8}{\percent} at least one location mention.
We mentioned earlier that cases exist where two or more action mentions share the same token span. \num{340} out of all \num{3563} annotated event mentions in \acs{fcct} fall into this category. A further \num{154} event mentions did not have a counterpart in the event database (such as matches from national football leagues). We jointly assigned these mentions to the coreference cluster \texttt{other\_event}.

By creating the \ac{fcct}, a reannotation and extension of \ac{fcc} on token level, we provide the first \ac{cdevcr} corpus featuring a large body of cross-subtopic event coreference links which is compatible with the existing \ac{ecbp} and \ac{gvc} corpora.\footnote{The \acf{fcct} is available at \corpusurl{}} This greatly expands the possibilities for \ac{cdevcr} research over multiple corpora, as we will demonstrate in \Crefrange{sec:exp_in_dataset}{sec:exp_cross_dataset}.

\section{Defining a General \ac{cdevcr} System}
\label{sec:system}
Recent \ac{cdevcr} approaches such as neural end-to-end systems or cluster-pair approaches were shown to offer great performance~\cite{barhom2019revisiting}, yet their black box nature and their complexity makes it difficult to analyze their decisions. In particular, our goal is to identify which aspects of a \ac{cdevcr} corpus are the strongest signals for event coreference which cannot be adequately investigated with recent \ac{cdevcr} systems. We therefore propose a conceptually simpler mention pair \ac{cdevcr} approach which uses a broad set of handcrafted features for resolving event coreference in different environments. We thus focus on developing an interpretable system, whereas reaching state-of-the-art performance is of secondary importance. This section explains the inner workings of the proposed system.

\subsection{Basic System Definition}
\label{sec:system_definition}
We resolve \acl{cdevc} by considering pairs of event mentions.
At training time, we sample a collection of training mention pairs. For each pair, we extract handcrafted features with which we train a probabilistic binary classifier that learns the coreference relation between a pair (\textit{coreferring} or \textit{not coreferring}). The classifier is followed by an agglomerative clustering step which uses each pair's coreference probability as the distance matrix.
At prediction time, all $\binom{n}{2}$ mention pairs are being classified without prior document preclustering.
For the reasons outlined in \Cref{sec:mention_detection_argument}, we choose to omit the mention detection step and work with the gold event mentions of each corpus throughout all experiments.

\subsection{Pair Generation for Training}
\label{sec:mention_pair_generation}
We explain three issues which arise when sampling training mention pairs and how we address them in our system.

The straightforward technique for sampling training pairs is to sample a fixed number of all possible coreferring and non-coreferring mention pairs. Due to the sparsity of the \ac{cdevcr} relation, the resulting set of pairs would mostly consist of non-coreferring pairs when using this technique, with the majority of coreferring pairs left unused. This issue has been partially addressed in the past with weighted sampling to increase the ratio of coreferring pairs~\cite{lee2012joint, barhom2019revisiting}.

We identified a second issue, namely the underrepresentation of mention pairs from the long tail, which weighted sampling does not address: We previously established that cluster sizes in corpora are imbalanced (see \Cref{fig:cluster_size_distribution}). If all $\binom{n}{2}$ coreferring pairs are generated for each cluster, the generated pairs will largely consist of pairs from the largest clusters.\footnote{For example, sorting all event clusters of the \ac{ecbp} training split by size, the largest cluster with 38 mentions would produce more coreferring pairs than the smallest \SI{50}{\percent} of clusters produce together.}
Manual inspection revealed that the variation in how events are expressed is limited, with large clusters exhibiting many action mentions with (near-)identical surface forms.\footnote{This is particularly pronounced for \ac{fcct} where events of football tournaments are mostly mentioned by \corpusexample{tournament} or \corpusexample{World Cup}.}
Consequentially, with common pair generation approaches, there is a high chance of generating many mention pairs which carry little information for the classifier, while mention pairs from clusters in the long tail are unlikely to be included.

Another issue we have not yet seen addressed in related work is the distribution of link types in the body of sampled pairs: in terms of the number of mention pair candidates available for sampling, the cross-topic link candidates strongly outnumber the cross-subtopic link candidates who in turn strongly outnumber the within-subtopic link candidates (and so on) by nature of combinatorics. This particularly concerns the large body of non-coreferring pairs. An underrepresentation of one of these types during training could cause deficiencies for the affected type at test time, hence care must be taken to achieve a balanced sampling.

We address these three issues as follows: (1)~We use the distribution of cluster sizes in the corpus to smoothly transition from generating all $\binom{n}{2}$ coreferring pairs for the smallest clusters to generating $(n-1) \cdot c$ pairs for the largest clusters, where $c \in \mathbb{R}^+$ is a hyperparameter. (2)~For each type of coreference link (within-document, within-subtopic, etc.) we sample up to $k$ non-coreferring mention pairs for each coreferring pair previously sampled for this type. Details on the sampling approach are provided in \Cref{sec:app_mention_pair_generation}.

\subsection{Features and Preprocessing}
\label{sec:feature_preprocessing}

Related work has demonstrated a great variety in the representations and features used to resolve cross-document event coreference (see \Cref{tab:feature_comparison_related_work} on \cpageref{tab:feature_comparison_related_work}), yet it remains unclear which features contribute the most to the coreference resolution performance on each of the three corpora. We therefore chose to implement a series of preprocessing steps and feature extractors which cover the majority of features used in previous systems.

\subsubsection{Preprocessing}
We perform lemmatization and temporal expression extraction with CoreNLP~\cite{manning2014stanford,chang2012sutime}, using document publication dates to ground temporal expressions for \ac{gvc} and \ac{fcct}.
We manually converted complex TIMEX expressions into date and time (so that \texttt{2020-01-01TEV} becomes \texttt{2020-01-01T19:00}).
For \ac{ecbp} and \ac{gvc} where participant, time, and location mentions are not linked to the action mention, we applied the SRL system by~\citet{shi2019simple} as implemented in AllenNLP~\cite{gardner2018allennlp}. We map spans with labels \texttt{ARGM-DIR} or \texttt{ARGM-LOC} to the location, \texttt{ARGM-TM} to the time and \texttt{ARG0} or \texttt{ARG1} to the participants of each respective event mention.
For all corpora we perform entity linking to DBPedia\footnote{\url{https://dbpedia.org} -- We used the \href{https://databus.dbpedia.org/dbpedia/collections/latest-core}{latest release} from April 1st, 2020.} via DBPedia Spotlight~\cite{mendes2011dbpedia}.

\subsubsection{Features}
The list of handcrafted mention pair features includes (1) string matching on action mention spans, (2) cosine similarity of TF--IDF vectors for various text regions, (3) the temporal distance between mentions, (4) the spatial distance between event actions based on DBPedia, and (5) multiple features comparing neural mention representations. These include representations of action mentions, embeddings of the surrounding sentence and embeddings of Wikidata entities which we obtained via the DBPedia entity linking step. Details on each feature are reported in \Cref{sec:app_feature_details}.

\subsection{Implementation Details}
We implemented the system using Scikit-learn~\cite{pedregosa2011scikit}. To obtain test predictions, we applied the following steps separately for each corpus:
We perform feature selection via recursive feature elimination~\cite{guyon2002gene} on the respective development split. We use a random forest classifier tasked with classifying mention pairs as "coreferring" / "not coreferring" as an auxiliary task for this stage.
We then identified the best classification algorithm to use as the probabilistic mention classifier. We tested logistic regression, a \acl{mlp}, a probabilistic SVM and XGBoost~\cite{chen2016xgboost}. We tuned the hyperparameters of each classifier via repeated 6-fold cross-validation for 24 hours on the respective training split.\footnote{Details on the hyperparameter optimization procedure are provided in \Cref{sec:app_hyperopt}.}.
Using the best classifier, we optimized the hyperparameters of the agglomerative clustering step (i.e., the linkage method, cluster criterion and threshold) for another 24 hours on the training split. For each experiment, we train five models with different random seeds to account for non-determinism. At test time we evaluate each of the five models and report the mean of each evaluation metric.\footnote{N.B. This applies to every result originating from this model throughout this work. We will therefore not point this out further.}

\section{Generalizability of \ac{cdevcr} Systems}
\label{sec:exp_in_dataset}

We train and test two \ac{cdevcr} systems and several baselines separately on the \ac{ecbp}, \ac{fcct}, and \ac{gvc} corpora to evaluate how flexibly these systems can be applied to different corpora (i.e., whether their overall design is sufficiently general for resolving \acl{cdevc} in each corpus).
The two systems are our proposed general system (see \Cref{sec:system}) and the system of \citet{barhom2019revisiting} (\acs{ba2019}). We chose \acs{ba2019} because it is the best-performing \ac{ecbp} system for which an implementation is available.

In \nameCrefs{sec:evaluation_metrics} \labelcref{sec:evaluation_metrics} and \labelcref{sec:baselines}, we define evaluation metrics and baselines. We then establish the performance of the feature-based system (\Cref{sec:in_dataset_feature_based_experiments}) on the three corpora, including a detailed link-level error analysis which we can only perform with this system. In \Cref{sec:exp_sota}, we explain how we apply \acs{ba2019} and compare its results to those of the feature-based system, analyzing the impact of document preclustering on the coreference resolution performance in the process.

\subsection{Evaluation Metrics}
\label{sec:evaluation_metrics}
Related work on \ac{cdevcr} has so far only scored predictions with the CoNLL F1~\cite{pradhan2014scoring} metric (and its constituent parts MUC~\cite{vilain1995model}, CEAF$_e$~\cite{luo2005on} and B$^3$~\cite{bagga1998entity}). We additionally score predictions with the LEA metric~\cite{moosavi2016which}. LEA is a link-based metric which, in contrast to other metrics, takes the size of coreference clusters into account. The metric penalizes incorrect merges between two large clusters more than incorrect merges of mentions from two singleton clusters. As we have shown that cluster sizes in \ac{cdevcr} corpora vary considerably (see \Cref{fig:cluster_size_distribution}) this is a particularly important property. LEA was furthermore shown to be more discriminative than the established metrics MUC, CEAF$_e$, B$^3$ and CoNLL F1~\cite{moosavi2016which}.

\subsection{Baselines}
\label{sec:baselines}
We report the two commonly chosen baselines \texttt{lemma} and \texttt{lemma-$\delta$} as well as a new \texttt{lemma-time} baseline based on temporal information:
\begin{enumerate}
	\item \texttt{lemma}: Action mentions with identical lemmas are placed in the same coreference cluster.
	\item \texttt{lemma-$\delta$}: Document clusters are created by applying agglomerative clustering with threshold $\delta$ on the TF--IDF vectors of all documents, then \texttt{lemma} is applied to each document cluster. For hyperparameter $\delta$, we choose the value which produces the best LEA F1 score on the training split.
	\item \texttt{lemma-time}: A variant of \texttt{lemma-delta} based on document-level temporal information. To obtain the time of the main event described by each document, we use the first occurring temporal expression or alternatively the publication date of each document. We create document clusters via agglomerative clustering where the distance between two documents is defined as the difference of their dates in hours. We then apply \texttt{lemma} to each document cluster. Here, the threshold $\delta$ represents a duration which is optimized as in \texttt{lemma-delta}.
\end{enumerate}

\subsection{Establishing the Feature-based System}
\label{sec:in_dataset_feature_based_experiments}
We run in-dataset experiments to determine the performance of the feature-based \ac{cdevcr} approach on each individual corpus. Details on the splits used for each corpus are reported in \Cref{sec:app_corpus_splits}.
When generating mention pairs for training, we undersample coreferring pairs (see \Cref{sec:mention_pair_generation}) using hyperparameters $c=8$ and $k=8$. In experiments involving \ac{fcct}, we use $c=2$ and $k=8$ to compensate for the large clusters in this corpus. Details on the choice of hyperparameters are provided in \Cref{sec:app_mention_pair_generation}. On all three corpora, the best mention pair classification results were obtained with XGBoost which led us to use it for all subsequent experiments with this system.

\subsubsection{Mention Clustering Results}
\label{sec:in_dataset_feature_based_clustering_results}
\begin{table}
	\centering
	\begin{tabular}{@{}ll*{4}{S[table-format=2.1, detect-weight=true]}@{}}
		\toprule
		Corpus                                 & System         & \multicolumn{1}{c}{CoNLL} &                  \multicolumn{3}{c}{LEA}                  \\
		\cmidrule(lr){3-3}
		\cmidrule(l){4-6} &                & {F1}                      & {P}               & {R}               & {F1}              \\
		\midrule
		\ac{ecbp}                              & lemma          & 61.9053                   & 42.8045           & 43.4982           & 43.1486           \\
		& lemma-$\delta$ & 74.3987                   & \bfseries 71.5428 & 53.6916           & \bfseries 61.3449 \\
		\addlinespace                          & feature-based  & \bfseries 74.8099         & 67.8531           & \bfseries 55.0546 & 60.7871           \\
		\midrule
		\ac{fcct}                              & lemma          & 42.9238                   & \bfseries 38.3587 & 19.8697           & 26.1788           \\
		& lemma-$\delta$ & 42.9238                   & \bfseries 38.3587 & 19.8697           & 26.1788           \\
		& lemma-time     & 39.7932                   & 36.8327           & 14.2433           & 20.5427           \\
		\addlinespace                          & feature-based  & \bfseries 54.2654         & 30.4189           & \bfseries 60.4073 & \bfseries 39.7645 \\
		\midrule
		\ac{gvc}                               & lemma          & 33.7975                   & 08.80542          & 29.7228           & 13.586            \\
		& lemma-$\delta$ & 50.2972                   & 43.8354           & 28.6605           & 34.6598           \\
		& lemma-time     & 51.4669                   & 53.813            & 27.2605           & 36.1887           \\
		\addlinespace                          & feature-based  & \bfseries 59.3876         & \bfseries 56.5137 & \bfseries 38.1787 & \bfseries 45.5702 \\
		\bottomrule
	\end{tabular}
	\caption{In-dataset \acs{cdevcr} results of baselines and the feature-based system. The full set of metrics is reported in \Cref{sec:app_extended_in_dataset_results}.}
	\label{tab:results_feature_baselines_in_dataset}
\end{table}

The results are shown in \Cref{tab:results_feature_baselines_in_dataset}. For brevity, we only report cross-document performance. It is obtained by applying the evaluation metrics on modified gold and key files in which all documents were merged into a single meta document~\cite{upadhyay2016revisiting}.

As was initially shown by \citet{upadhyay2016revisiting}, the \texttt{lemma-$\delta$} baseline is a strong baseline on the \ac{ecbp} corpus. The feature-based system performs on par with this baseline.

For \ac{fcct}, the optimal $\delta$ produces a single cluster of all documents which leads to identical results for the \texttt{lemma} and \texttt{lemma-$\delta$} baselines. This is a direct consequence of the fact that in this corpus, the majority of event coreference links connect documents from different subtopics. In contrast to \ac{ecbp}, where preclustering documents by textual content produces document clusters which are near-identical to the gold subtopics~\cite{barhom2019revisiting, cremisini2020new}, such a strategy is disadvantageous for \ac{fcct} because the majority of coreference links would be irretrievably lost after the document clustering step.
The \texttt{lemma-time} baseline performs worse on \ac{fcct} than \texttt{lemma-$\delta$}, indicating that the document publication date is less important than the document content. The feature-based approach outperforms the baselines on \ac{fcct}, showing higher recall but lower precision which indicates a tendency to overmerge clusters.

The lemma baselines perform worse on \ac{gvc} than on \ac{ecbp} in absolute numbers which can be attributed to the fact that \citet{vossen2018dont} specifically intended to create a corpus with ambiguous event mentions. Furthermore, the baseline results show that knowing about a document's publication date is worth more than knowing its textual content (at least for this corpus). The feature-based system mostly improves over the baselines in terms of recall.

Another noteworthy aspect in \Cref{tab:results_feature_baselines_in_dataset} are the score differences between CoNLL F1 and LEA F1. In the within-document entity coreference evaluations performed by \citet{moosavi2016which} alongside the introduction of the LEA metric, the maximum difference observed between CoNLL F1 and LEA F1 were roughly \SI{10}{\pp}. Our experiments exhibit differences of \SI{14}{\pp} for systems and up to \SI{20}{\pp} for baselines due to imbalanced cluster sizes in \ac{cdevcr} corpora.

\subsubsection{Mention Pair Classifier Results}
\label{sec:exp_in_dataset_classifier}

\begin{table*}
	\small
	\centering
	\setlength{\tabcolsep}{4.25pt}
	\begin{tabular}{@{}lS[table-format=2.0, table-space-text-post=$\:$\si{\kilo}]*{3}{S[table-format=2.1, detect-weight=true]}S[table-format=3.0, table-space-text-post=$\:$\si{\kilo}]*{3}{S[table-format=2.1, detect-weight=true]}S[table-format=3.0, table-space-text-post=$\:$\si{\kilo}]*{3}{S[table-format=2.1, detect-weight=true]}@{}}
		\toprule
		Link type                                                       &                          \multicolumn{4}{c}{\acs{ecbp}}                          &                          \multicolumn{4}{c}{\acs{fcct}}                           &                           \multicolumn{4}{c}{\acs{gvc}}                            \\
		\cmidrule(lr){2-5}
		\cmidrule(lr){6-9}
		\cmidrule(l){10-13} & {Links}              & {P}               & {R}               & {F1}              & {Links}               & {P}               & {R}               & {F1}              & {Links}               & {P}               & {R}               & {F1}               \\ \midrule
		within-document                                                 & 10.751$\:$\si{\kilo} & 57.63358180298196 & 55.85321100917431 & 56.69823481284595 & 6.567$\:$\si{\kilo}   & 53.49548264585289 & 56.16022099447514 & 54.79047499598457 & 7.239$\:$\si{\kilo}   & 69.13548924090189 & 30.59912854030501 & 42.421888648205625 \\
		within-subtopic                                                 & 83.191$\:$\si{\kilo} & 64.99720446671013 & 54.51134930643128 & 59.29403250822901 & 21.481$\:$\si{\kilo}  & 51.9978520070693  & 48.44322344322344 & 50.15536414267515 & 9.650$\:$\si{\kilo}   & 70.93470290594713 & 28.86924150767799 & 41.03435996839956  \\
		cross-subtopic                                                  & 86.805$\:$\si{\kilo} & 0.0               & 0.0               & 0.0               & 518.486$\:$\si{\kilo} & 53.73427505892876 & 36.27549169025329 & 43.30336552779629 & 434.836$\:$\si{\kilo} & {n/a}             & {n/a}             & {n/a}              \\ \bottomrule
	\end{tabular}
	\caption{Mention pair classifier performance of the feature-based system for each cross-document coreference link type. "Coreferring" is used as the positive class. The "Links" column shows the total number of links (coreferring and non-coreferring) per type and corpus based on which P/R/F1 were calculated.}
	\label{tab:results_feature_baselines_classifier_in_dataset}
\end{table*}

To evaluate the probabilistic mention pair classifier in isolation for different corpora and coreference link types, we compute binarized recall, precision and F1 with respect to gold mention pairs.\footnote{Note that this approach puts higher weight on large clusters, as these produce a greater number of mention pairs. It is nonetheless the only evaluation approach we are aware of which permits analyzing performance \textit{per link type}. Link-based coreference metrics such as MUC~\cite{vilain1995model} cannot be used as a replacement, as these (1) require a full clustering opposed to one score per pair and (2) by design abstract away from individual links in a system's response.}
The results are reported in \Cref{tab:results_feature_baselines_classifier_in_dataset}.
The \ac{gvc} test split does not contain coreferring cross-subtopic event coreference links, therefore these cells are marked with "n/a". Five links of this type are present in the \ac{ecbp} test split of which none were resolved by the in-dataset \ac{ecbp} model.
For \ac{fcct} and \ac{gvc}, the performance in resolving within-document, within-subtopic and cross-subtopic event coreference links decreases gradually from link type to link type. This suggests that the greater the distance covered by an event coreference link is in terms of the topic-subtopic-document hierarchy of a corpus, the more difficult it becomes to resolve it correctly.

\subsubsection{Error Analysis}
\label{sec:in_dataset_feature_based_classifier_error_analysis}

\begin{table*}
	\centering
	\small
	\begin{tabular}{@{}rp{.45\textwidth}p{.45\textwidth}@{}}
		\toprule
		Ex.	& Mention context A	& Mention context B \\
		\midrule
		1	& \multicolumn{2}{l}{\textbf{\acs{fcct}, cross-subtopic link, false positive}} \\
		\addlinespace
		& Zlatko Dalic's men will be back at the Luzhniki Stadium on Sunday to face France, who glanced a 1--0 \textbf{victory} over Belgium on Tuesday thanks to a Samuel Umtiti header.	& Belgium claims third place with a 2--0 \textbf{win} over England \\
		\midrule
		2	& \multicolumn{2}{l}{\textbf{\acs{gvc}, cross-subtopic link, false positive}} \\
		\addlinespace
		& A 66-year-old man was shot in his leg and \textbf{grazed} in his arm early Monday while sitting on a park bench in Charles Village, police said.	& The victims -- except an 18-year- old man who refused medical attention after a bullet \textbf{grazed} his left leg -- were taken to UAMS Medical Center. \\
		\midrule
		3	& \multicolumn{2}{l}{\textbf{\acs{ecbp}, cross-subtopic link, false positive}} \\
		\addlinespace
		& Tuesday, July 2, 2013. A strong \textbf{earthquake} struck Indonesia's Aceh province on Tuesday, killing at least one person and leaving two others missing.	& THOUSANDS of frightened residents flooded makeshift refugee camps in Indonesia's West Papua province today after two powerful \textbf{earthquakes} flattened buildings and killed at least one person. \\
		\bottomrule
	\end{tabular}
	\caption{Notable misclassifications found during manual error analysis}
	\label{tab:error_analysis_examples}
\end{table*}

To gain a better understanding of the system's limitations, we manually analyzed predictions of the mention pair classifier. We analyzed five false positive and five false negative cases for each link type and corpus (roughly 90 mention pairs in total).

We found that textual similarity between action mentions accounts for a large portion of mistakes on the \ac{ecbp} and \ac{gvc} corpora: unrelated but similar action mentions caused false positive cases and vice-versa, coreferring but merely synonymous action mentions led to false negative cases.
The \ac{fcct} model did not resolve coreference well between mentions like \corpusexample{the tournament}, \corpusexample{this year's cup} and \corpusexample{2018 World Cup}, contributing to false negative cases. Also, the model showed a tendency of merging event mentions prematurely when the action mention and at least one participant matched (see example 1 in \Cref{tab:error_analysis_examples}) which would explain the high recall and low precision results seen in \Cref{tab:results_feature_baselines_in_dataset}.
For all three models, we noticed misclassifications when a sentence contained multiple event mentions (see example 2 in \Cref{tab:error_analysis_examples}). In the given example, it is likely that information from the unrelated \corpusexample{shot in his leg} mention leaked into the representation of the \corpusexample{grazed} event which contributed to the incorrect classification.
For \ac{ecbp}, we noticed that the lack of document publication date information makes certain decisions considerably harder. For example, the earthquake events seen in example 3 are unrelated and took place four years apart. While one could come to this conclusion with geographic knowledge alone (the provinces lie on opposite sides of Indonesia), date information would have made this decision easier.

It is reassuring that many of the shortcomings we found would be fixable with a cluster-level coreference resolution approach, (joint) resolution of entity coreference, injection of corpus-specific world knowledge (a football match must take place between exactly two teams, etc.) or with annotation-specific knowledge (for example knowledge of \citet{vossen2018dont}'s domain model for the annotation of \ac{gvc}). Our system could be improved by incorporating these aspects, however at the cost of becoming more corpus-specific and less interpretable.

\subsection[Comparison to Barhom et al. 2019]{Comparison to \citet{barhom2019revisiting}}
\label{sec:exp_sota}
We test an established \ac{cdevcr} system, the former state-of-the-art neural \ac{cdevcr} approach of \citet{barhom2019revisiting} (\acs{ba2019}), for its generalization capabilities on the three corpora.

\subsubsection{Experiment Setup}
We trained one model of \acs{ba2019} for each corpus. \acs{ba2019} can resolve event and entity coreference jointly. For the sake of comparability we only use the event coreference component of this system for all experiments since \ac{fcct} and \ac{gvc} do not contain entity coreference annotations. We replicate the exact data preprocessing steps originally used for \ac{ecbp} on \ac{fcct} and \ac{gvc}. This includes the prediction of semantic roles with the SwiRL \ac{srl} system~\cite{surdeanu2007combination}.
The \ac{fcct} corpus mainly consists of cross-subtopic event coreference links (see \Cref{sec:annotation_results}). The trainable part of the \acs{ba2019} system (mention representation and agglomerative mention clustering) is meant to be trained separately on each subtopic of a corpus. This is because at prediction time, the partitioning of documents into subtopics will already be handled by a foregoing and separate document preclustering step. In order not to put \acs{ba2019} at a disadvantage for \ac{fcct}, we train it on three large groups of documents which correspond to the three football tournaments present in the \ac{fcct} training split instead of training it on the actual \ac{fcct} subtopics. For this corpus, we also apply undersampling with the same parameters as for the feature-based system.

\subsubsection{Options for Document Preclustering}
\label{sec:options_document_clustering}
Clustering documents by their content before applying mention-level event coreference boosts performance on the \ac{ecbp} corpus~\cite{cremisini2020new, barhom2019revisiting, choubey2017event, upadhyay2016revisiting}. As was recommended by \citet{cremisini2020new}, we report results for several document preclustering strategies to better distinguish the source of performance gains or losses.
We compare (1) no preclustering, (2) the gold document clusters, and (3) the k-means clustering approach used in \cite{barhom2019revisiting}.
The gold document clusters are defined via the transitive closure of all event coreference links.
For \ac{gvc}, this gold document clustering is identical to the corpus subtopics. For the \ac{fcct} test split, the gold clustering is a single cluster containing all documents which is equivalent to not applying document clustering at all.
For \ac{ecbp}, the gold clustering largely corresponds to the corpus subtopics with the exception of some subtopics which are merged due to cross-subtopic event coreference links.
In the k-means approach, all $n$ input documents are represented by TF--IDF vectors based on which all possible k-means clusterings for $k=2,\ldots, n$ are created.
From these clusterings, the one with the highest silhouette score~\cite{rousseeuw1987silhouettes} is used.

\subsubsection{Results}
\label{sec:exp_sota_results}

\begin{table*}
	\centering
	\begin{tabular}{@{}lll*{4}{S[table-format=2.1, detect-weight=true]}@{}}
		\toprule
		Corpus                                 & System       & Preclustering       & \multicolumn{1}{c}{CoNLL} &   \multicolumn{3}{c}{LEA}   \\
		\cmidrule(lr){4-4}
		\cmidrule(l){5-7} &              &                     & {F1}                      & {P}     & {R}     & {F1}    \\
		\midrule
		\ac{ecbp}                              & ours         & none $\bullet$      & 74.8099                   & 67.8531 & 55.0546 & 60.7871 \\
		&              & k-means             & 75.3664                   & 68.9013 & 55.6042 & 61.5423 \\
		&              & gold                & 76.2525                   & 71.5698 & 56.3787 & 63.0718 \\
		\addlinespace                          & \acs{ba2019} & k-means             & 79.21                     & 67.17   & 68.04   & 67.60   \\
		&              & gold                & 79.08                     & 66.67   & 68.50   & 67.57   \\
		\midrule
		\acs{fcct}                             & ours         & none/gold $\bullet$ & 54.2654                   & 30.4189 & 60.4073 & 39.7645 \\
		\addlinespace                          & \acs{ba2019} & none/gold           & 48.05                     & 39.94   & 27.18   & 32.35   \\
		&              & k-means             & 34.61                     & 36.93   & 08.56   & 13.90   \\
		\midrule
		\ac{gvc}                               & ours         & none $\bullet$      & 59.3876                   & 56.5137 & 38.1787 & 45.5702 \\
		&              & k-means             & 60.7694                   & 63.866  & 38.5545 & 48.0825 \\
		&              & gold                & 62.8363                   & 66.1115 & 41.0781 & 50.6714 \\
		\addlinespace                          & \acs{ba2019} & k-means             & 73.26                     & 58.68   & 74.31   & 65.58   \\
		&              & gold                & 79.80                     & 69.49   & 80.89   & 74.76   \\
		\bottomrule
	\end{tabular}
	\caption{Comparison of \ac{cdevcr} performance depending on preclustering strategy. All results reported use gold event mentions. Results marked with $\bullet$ are taken from \Cref{tab:results_feature_baselines_in_dataset}.}
	\label{tab:results_sota}
\end{table*}

Due to long runtimes times of \acs{ba2019}\footnote{Training and optimizing one model on the \ac{fcct} corpus took 10 days on an Nvidia V100 GPU.} results reported from this system stem only from a single execution. We do not report experiments without preclustering on \ac{ecbp} and \ac{gvc} using this system due to scalability issues caused by the greater number of event mentions in these corpora (see \Cref{tab:dataset_properties} on \cpageref{tab:dataset_properties}). The results are shown in \Cref{tab:results_sota}.

We comment on the most remarkable results. The \acs{ba2019} system architecture performs well on \ac{gvc}, reaching 65.6 LEA F1. Compared to the \ac{ecbp} results, there is a notable score difference between the k-means and gold preclustering variants on this corpus.
The reason is the same one that led to the \texttt{lemma-time} baseline outperforming \texttt{lemma-$\delta$} on this corpus --- preclustering documents by textual content is less effective on a corpus with a single topic, and \acs{ba2019} does not make use of document publication date annotations.
For \ac{fcct}, applying \acs{ba2019} out-of-the-box with k-means preclustering performs much worse than when the preclustering step is omitted due to the large amount of cross-subtopic links which are being cut off.

When comparing systems against each other, \acs{ba2019} performs better than the feature-based approach on \ac{ecbp} and \ac{gvc}. The opposite is the case for \ac{fcct} where the neural model shows greatly reduced recall in comparison to the feature model.
This is surprising to some extent since \acs{ba2019} is a more powerful cluster-level approach compared to the mention pair approach.
A plausible explanation for the performance drop on \ac{fcct} is the narrower set of features in \acs{ba2019}. Notably, this system lacks world knowledge on locations and participants and does not explicitly model temporal information, all of which would make intuitive sense to have for a corpus mentioning a variety of football players and matches happening on specific dates.
The next section adds evidence to this intuition by analyzing in greater depth the information necessary for resolving event coreference in each corpus.

With respect to the experiments conducted in this section, we have shown that it cannot be taken for granted that \ac{cdevcr} systems are sufficiently general to perform equally well on different corpora.
This concerns both the quality of their results (which can fluctuate) as well as more fundamental aspects such as their computational complexity (which may preclude their applicability). In the concrete case of \acs{ba2019}, this comes down to the choice of the input features and the dependency on document preclustering.

\section{Identifying the Signals for Event Coreference}
\label{sec:exp_signals}
According to the \ac{cdevcr} task definition (see \Cref{sec:cdevcr_task_definition}), coreference between a pair of event mentions requires a match between each of their components (action, participants, time, location).
We analyze to which extent corpora satisfy this definition in practice, namely, whether inference over all event components is indeed required, or whether certain event components suffice as signals for resolving event coreference.
We approach this analysis in two ways:
(1) We investigate the most important features per corpus at training time via model introspection, and (2) we mask the mentions of certain event components in the test split and measure the impact on test performance.
We explain the two approaches and present their results (\Cref{sec:feature_importance,sec:masking}), then jointly discuss their outcome in \Cref{sec:signals_summary}.

\subsection{Feature Importance}
\label{sec:feature_importance}

Our main reason for developing a feature-based system was that, compared to neural systems, it allows one to directly analyze which input information a model is making use of.

In our system architecture, the agglomerative clustering step is preceded by a mention pair classifier which we found worked best with the decision tree boosting framework XGBoost (see \Cref{sec:exp_in_dataset}). For decision trees, feature importance metrics can be derived from trained models. In \Cref{tab:feature_importance_top}, we report the top features selected during feature selection for each corpus. Alongside, we report the importance of each feature at training time according to the \textit{gain} metric of XGBoost.\footnote{Gain refers to the gain in accuracy from introducing a split in a decision tree using a particular feature. See \url{https://xgboost.readthedocs.io/en/latest/tutorials/model.html} for more details.}

\begin{table*}
	\centering
	\setlength{\tabcolsep}{3pt}
	\begin{tabular}{@{}S[table-format=0.3]ll@{}}
		\toprule
		{Gain}			& Feature Name							& Short description \\
		\midrule
		\multicolumn{3}{@{}l}{\textbf{\acs{ecbp}}} \\
		0.342106		& is-lemma-identical					& Lemma identity between actions \\
		0.248146		& surface-form-mlipns-distance			& MLIPNS distance on action surface forms \\
		0.130495		& action-mention						& SpanBERT action similarity \\
		0.0850028		& is-surface-form-identical				& Action surface form identity \\
		0.0555572		& document-similarity					& TF--IDF document similarity \\
		\midrule
		\multicolumn{3}{@{}l}{\textbf{\acs{fcct}}} \\
		0.301784		& surface-form-mlipns-distance			& MLIPNS distance on action surface forms \\
		0.144558		& is-lemma-identical					& Lemma identity between actions \\
		0.0983997		& distance-closest-overall-level-year	& Years between closest temporal expressions \\
		0.0805005		& is-surface-form-identical				& Surface form identity between actions \\
		0.0534523		& distance-sentence-level-year			& Years between temp. exprs. in same sentence \\
		\midrule
		\multicolumn{3}{@{}l}{\textbf{\acs{gvc}}} \\
		0.254793		& surface-form-mlipns-distance			& MLIPNS distance on action surface forms \\
		0.126396		& distance-doc.-pub.-level-week			& Week distance of doc. publication dates \\
		0.125259		& distance-doc.-pub.-level-month		& Month distance of doc. publication dates \\
		0.075458		& document-similarity					& TF--IDF document similarity \\
		0.0593222		& is-lemma-identical					& Lemma identity between actions \\
		\bottomrule
	\end{tabular}
	\caption{The top five selected features per corpus with feature importance (see \Cref{sec:exp_in_dataset}). See \Cref{sec:app_feature_details} for detailed feature descriptions and \Cref{sec:app_feature_importance} for the full listing.}
	\label{tab:feature_importance_top}
\end{table*}

For \ac{ecbp}, the selected features only cover event actions and context representations. Few features were selected overall.
For \ac{fcct}, event action and temporal information received the greatest attention. There is a notable absence of document-level features which we attribute to the fact that document similarity is not of prime importance for resolving cross-subtopic coreference links.
For \ac{gvc} a large number of features was selected, suggesting that diverse information is required for coreference resolution decisions. The most prominent features cover the event action and document-level information.

\subsection{Masking of Event Components}
\label{sec:masking}
We want to analyze the impact of each type of event component at test time. To do so, we create variants of the test data where the spans of certain event mentions are masked. We then predict with the models trained in the in-dataset scenario (\Cref{sec:exp_in_dataset}) and measure the score delta.

When masking mention spans, we replace each token with a unique dummy token.\footnote{We use fixed-length random tokens from the set \texttt{[a-zA-Z]*{5}} .} This is to ensure that the string similarity between two mentions is entirely random. For action components, we replace all gold annotated mention spans. For participant, time, and location components, we replace gold annotations as well as any additional entities identified by DBpedia Spotlight. Masked spans are also removed from semantic role arguments. For the \ac{fcct} and \ac{gvc} corpora, we additionally mask the document publication date. This masking approach is not without limitations. In \ac{fcct} and \ac{gvc}, only a subset of all events was annotated. For participants and actions, the three corpora annotate only the head of the phrase. Both these cases may lead to information-bearing tokens leaking into the masked dataset. We nevertheless believe that our approach is an effective approach for analyzing the impact of specific event components at test time.

\begin{table*}
	\centering
	\includegraphics{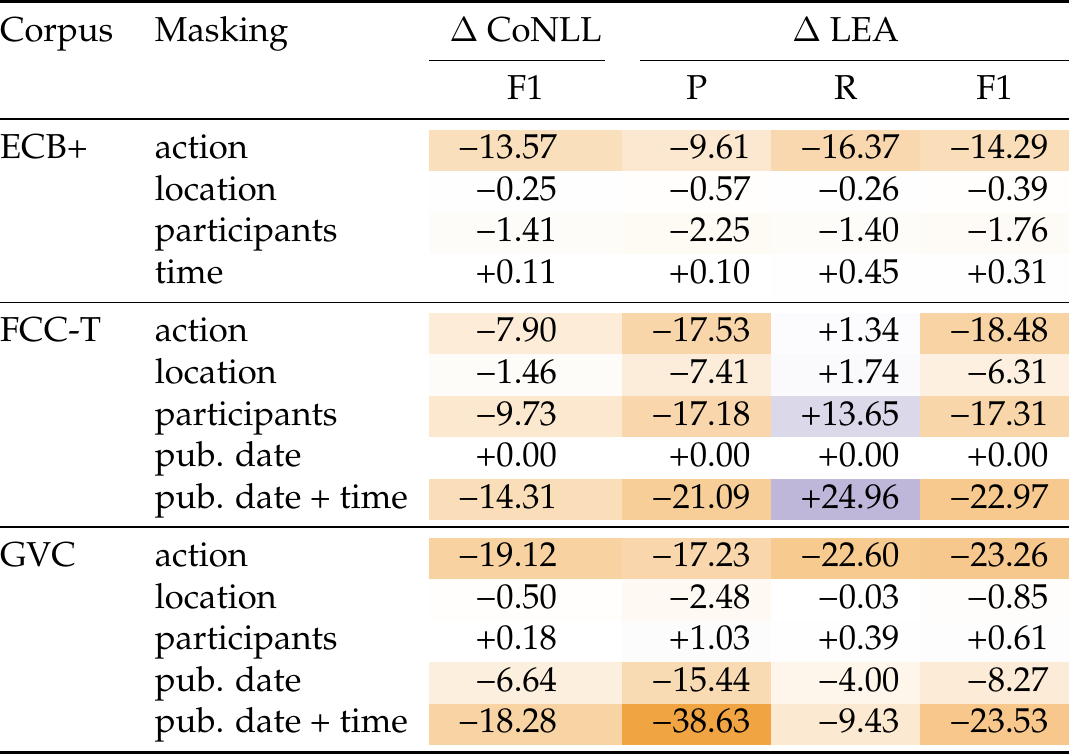}
	\caption{Impact on test performance when masking spans of certain event components. We report the score deltas of the feature-based system w.r.t. the scores from \Cref{tab:results_feature_baselines_in_dataset}.}
	\label{tab:analysis_event_components}
\end{table*}

The results are shown in \Cref{tab:analysis_event_components}.
On the \ac{ecbp} corpus, masking event actions has the strongest impact on performance. This is to be expected since the majority of features used by the model are action-related features.
For \ac{fcct}, masking intensifies the issue of cluster overmerging. Between action, participants and time, the drop in LEA F1 is comparable. When interpreting the effects of masking on \ac{fcct}, it is important to keep in mind that all events annotated in this corpus are \textit{planned} events whose time, location and (to some extent) participants are known in advance. This increases the frequency by which these event components are mentioned in text, and on the flipside should cause stronger losses in performance compared to \ac{ecbp} and \ac{gvc} which contain a smaller proportion of planned events. The fact that scores drop for \ac{fcct} only when the document publication date \textit{and} temporal expressions are removed indicates that the document publication date was not used for grounding temporal expressions in text.
In terms of the \ac{gvc}, action and time stand out as the event components with the highest impact whereas location and participant information barely contribute to the results. We were surprised to see that with regard to temporal information, temporal expressions in text carry more information than the document publication date. Manual inspection revealed that in sentences like \corpusexample{Two-year-old child shot in the chest in Palm Harbor}, the entity linker would frequently misclassify \corpusexample{Two-year-old} as a temporal expression instead of a person entity. A portion of the performance loss from masking time expressions may therefore arise from masked participants.

\subsection{Summary on the Signals for Event Coreference}
\label{sec:signals_summary}
Answering our initial question, whether established \ac{cdevcr} corpora match the \ac{cdevcr} task definition in that they require inference over each event component, we conclude that this is not the case.

Our experiments demonstrate that \acl{cdevcr} decisions in \ac{ecbp} are strongly driven by action mentions. \ac{gvc}, designed to overcome this shortcoming of \ac{ecbp}~\cite{vossen2018dont}, necessitates inference over action mentions, time and to some level participants, but does not challenge systems on spatial inference. \ac{fcct} is the most balanced of the three corpora based on the facts that at training time, feature selection yielded a broad selection of features and at test time, performance decreases similarly when action, location or time information is removed.
Overall, neither corpus requires inference over all four event components which define the \acl{cdevc} relation.

This indicates that \ac{cdevcr} systems which focus on solving a single corpus model only a subset of the entire \ac{cdevcr} task which severely limits their downstream use on unseen data, as this data may reflect the task differently from what was observed at training time.
The findings further raise the question to which degree it is possible to resolve \acl{cdevc} in the three corpora with a \textit{single} model, as this would be the most desirable usage scenario for applying \ac{cdevcr} downstream. We address this question in the following section.

\section{Generalizability of Trained \ac{cdevcr} Models}
\label{sec:exp_cross_dataset}

All preceding experiments in this work have addressed the \ac{ecbp}, \ac{fcct}, and \ac{gvc} corpora in isolation, training a separate model per corpus. In downstream application scenarios, such a differentation is not possible -- here, a \ac{cdevcr} system is expected to resolve \acl{cdevc} in a robust manner regardless of the selection of topics or underlying structure of a given collection of documents.
To gain insights into which performance to expect in such a scenario, we test models of the feature-based system in a \textit{cross-dataset} transfer scenario on unseen corpora.\footnote{We did not test \acs{ba2019} in this scenario due to the scalability issues reported in \Cref{sec:exp_sota}.}

Furthermore, recent research on the \acf{qa} task has shown that training systems jointly on multiple datasets can improve model robustness and can boost performance~\cite{talmor2019multiqa, guo2020multireqa, fisch2019mrqa}.
In this work, we have established compatibility between the \ac{ecbp}, \ac{fcct}, and \ac{gvc} corpora and have identified the different ways in which each corpus models event coreference. We test whether benefits similar to those observed for \ac{qa} are possible for \ac{cdevcr} by training the feature-based system on \textit{multiple} \ac{cdevcr} corpora.

\subsection{Experiment Setup}
We use same splits for all corpora as in previous experiments. In \ac{ecbp}, a number of topics cover sports news or news related to gun violence. We refer to these corpus subsets as \acs{ecbp}$_\text{sports}$ and \acs{ecbp}$_\text{guns}$ respectively and treat them separately in our experiments. Their contents are shown in \Cref{tab:ecbp_extra_splits}. The \ac{ecbp} test split remains unchanged.

\begin{table}
	\centering
	\begin{tabular}{@{}lll@{}}
		\toprule
		Corpus Subset	& Train	& Dev \\
		\midrule
		\acs{ecbp}$_\text{sports}$	& 5, 7, 10, 25	& 29, 31 \\
		\acs{ecbp}$_\text{guns}$	& 3, 8, 16, 18, 22	& 33 \\
		\bottomrule
	\end{tabular}
	\caption{Topics used as distinct subsets of \acs{ecbp}.}
	\label{tab:ecbp_extra_splits}
\end{table}

When combining two corpora, we use the union of features previously selected during feature selection. The increase in training data leads to an increase in mention pairs which prolongs the training process. We therefore optimize the hyperparameters of the mention pair classifier and the agglomerative clustering step for three days each.

\subsection{Results}

\begin{table*}
	\small
	\centering
	\setlength{\tabcolsep}{4.35pt}
	\begin{tabular}{@{}*{5}{c}*{12}{S[table-format=2.1, detect-weight=true]}@{}}
		\toprule
		\multicolumn{5}{c}{Train + Dev Split}	& \multicolumn{12}{c}{Test Split Performance (LEA)} \\
		\cmidrule{1-5}
		\cmidrule(l){6-17}
		\multirow[b]{3}{*}{\rotatebox[origin=c]{90}{\parbox[c]{1cm}{\acs{ecbp}}}}	& \multirow[b]{3}{*}{\rotatebox[origin=c]{90}{\parbox[c]{1cm}{\acs{ecbp}$_\text{sports}$}}}	& \multirow[b]{3}{*}{\rotatebox[origin=c]{90}{\parbox[c]{1cm}{\acs{ecbp}$_\text{guns}$}}}	& \multirow[b]{3}{*}{\rotatebox[origin=c]{90}{\parbox[c]{1cm}{\acs{fcct}}}}	& \multirow[b]{3}{*}{\rotatebox[origin=c]{90}{\parbox[c]{1cm}{\acs{gvc}}}} &	&	&	&	&	&	&	&	&	&	&	& \\
		\addlinespace
		&	&	&	&	& \multicolumn{3}{c}{\acs{ecbp}}							& \multicolumn{3}{c}{\acs{fcct}}							& \multicolumn{3}{c}{\ac{gvc}}								& \multicolumn{3}{c}{Harmonic Means} \\
		\cmidrule(lr){6-8}
		\cmidrule(lr){9-11}
		\cmidrule(lr){12-14}
		\cmidrule(l){15-17}
		&												&											&								&							& {P}	& {R}	& {F1}		& {P}	& {R}	& {F1}		& {P}	& {R}	& {F1}		& {P}	& {R}	& {F1} \\
		\midrule
		\checkmark	& \plhdr			& \plhdr			& \plhdr			& \plhdr			& \bfseries 67.85305954968877	& 55.05455494947548			& \bfseries 60.787136826441	& 41.42023397622781			& 08.524460588440183			& 14.134440954440702			& 32.096465241209404			& 25.26161301075749			& 28.27132251891441			& 42.83474794276155			& 17.137076477727678			& 24.4754995006248 \\
		\plhdr		& \plhdr				& \plhdr				& \checkmark	& \plhdr				& 15.96131731569325			& \bfseries 57.15711386994929	& 24.907385103300675			& 30.41894024174348			& \bfseries 60.40733495568591	& 39.76453832193483			& 03.5526492442660144			& \bfseries 74.9774494149505	& 06.781433642264527			& 07.957437104904444			& \bfseries 63.30815558751782	& 14.100513163555067 \\
		\plhdr		& \plhdr				& \plhdr				& \plhdr		& \checkmark	& 48.84504909141456			& 50.16833842982555			& 49.49658000370589			& 42.16430166186827			& 17.65703260181437			& 24.8543418799426				& \bfseries 56.51366580128873	& 38.17872866957683			& \bfseries 45.57023651600609	& \bfseries 48.4774177469631	& 29.194198092313606			& 36.4157295959342 \\
		\midrule
		\plhdr		& \checkmark	& \plhdr			& \checkmark	& \plhdr			& 52.06447308722967			& 54.27976093962945			& 53.13829704160747			& 42.71918700291826			& 40.24933816165808			& \bfseries 41.31191612305315	& 16.892331067148966			& 26.61796912156743			& 20.64424301755212			& 29.46546660620906			& 37.111806665143793			& 32.799587602450053 \\
		\plhdr		& \plhdr		& \checkmark	& \plhdr			& \checkmark	& 62.54681323670088			& 55.16719422923655			& 58.62544027960724			& 41.478215050120704			& 19.94434569437878			& 26.93465376528754			& 45.016429570795546			& 35.44894706146014			& 39.66339778359211			& 48.14545481149184			& 31.095874114862887			& 37.785042985887357 \\
		\checkmark	& \plhdr		& \plhdr			& \checkmark	& \plhdr			& 63.25672317658109			& 60.701338697687			& 58.03288627726739			& \bfseries 50.92106676240623	& 19.69907844434076			& 28.40139622328898			& 29.65042353403821			& 26.674242869041537			& 28.07149001909567			& 43.36941946660694			& 28.06166292005017			& 34.06594912372894 \\
		\checkmark	& \plhdr		& \plhdr			& \plhdr			& \checkmark	& 55.37047479809182			& 54.77442546022191			& 55.07036420021515			& 38.00609962253095			& 25.46761266709582			& 30.49346826332583			& 27.359811431603875			& 32.83516539986092			& 29.84021388543425			& 37.07287774367798			& 34.09966420865055			& 35.51795624320908 \\
		\plhdr		& \plhdr		& \plhdr			& \checkmark	& \checkmark	& 37.56206667064297			& 47.74465854529305			& 42.03851253552827			& 34.10112696249451			& 48.51992369849193			& 40.04400288717195			& 45.69485388518007			& 36.55810043515443			& 40.61763923438674			& 38.54483001327705			& 43.53595531057695			& \bfseries 40.883026960062213 \\
		\checkmark	& \plhdr		& \plhdr			& \checkmark	& \checkmark	& 24.28429177381188			& 53.595104615098				& 33.411812635837485			& 26.49204062441818			& 58.80177695888781			& 36.52164036386481			& 32.313692176180625			& 38.26419839500644			& 35.03191382326289			& 27.304325075180785			& 48.5446325720302				& 34.94224465471178 \\
		\bottomrule
	\end{tabular}
	\caption{Performance of the feature-based system when trained on a single respective corpus (top) vs. multiple corpora at once (bottom). No document preclustering was applied. The rightmost set of columns shows P, R and F1 aggregated over the three corpora. The full set of metrics is reported in \Cref{sec:app_full_cross_dataset_results}.}
	\label{tab:results_feature_cross_dataset}
\end{table*}

The results of our experiments, evaluated with LEA, are shown in \Cref{tab:results_feature_cross_dataset}.
As we have shown in preceding sections, the requirements for resolving \acl{cdevc} vary between the corpora which strongly influences the feature selection and model training processes. We hence expected models trained on a single corpus to perform poorly when evaluated on unseen corpora. This is confirmed by the top rows of \Cref{tab:results_feature_cross_dataset} where significant gaps between in-dataset and cross-dataset performance can be observed.

When looking at the performance \textbf{on individual corpora}, models trained on multiple corpora perform consistently worse than those trained on a single corpus, with few exceptions (mixing \ac{fcct} with  \acs{ecbp}$_\text{sports}$ or \ac{gvc} during training leads to more balanced LEA recall and precision).
However, the best overall result in terms of the whole task (i.e., \textbf{across all corpora}) was achieved with joint training: the model trained on \ac{fcct} and \ac{gvc} scores \num{40.9} mean LEA F1 over all corpora, whereas the best single-corpus model trained on \ac{gvc} alone only reached \num{36.4}.
In conclusion, training on multiple corpora did not boost performance on individual corpora. Nevertheless, joint training on multiple corpora has emerged as an important strategy for reaching general \ac{cdevcr} systems. 

We have only scratched the surface of joint training for \ac{cdevcr}. Further improvements may be achieved with more sophisticated training approaches, for example by mixing together different amounts of each corpus (potentially aiming for certain distributions of coreference link types), testing its effects on \ac{cdevcr} systems beyond mention pair approaches or performing training data augmentation with data from other NLP tasks.

\section{Discussion}
\label{sec:discussion}
Despite its importance for downstream applications, the generalizability of \ac{cdevcr} systems over different corpora has not received attention in the past.

Our experiments showed that a system achieving state-of-the-art-level performance on \ac{ecbp} does not consistently produce results of the same quality when trained and tested on other \ac{cdevcr} corpora (see \Cref{sec:exp_sota}).
This raises the suspicion that similar systems which were developed for a single corpus lack the capacity of generalizing to unseen corpora. This suspicion is substantiated when looking at the results of our cross-dataset experiments. We showed that training a general, feature-based \ac{cdevcr} system on a single corpus yields good results on the test split of that respective corpus, whereas performance on other corpora falls short of these results (see \Cref{sec:exp_cross_dataset}).

This is due to the fact that the \ac{ecbp}, \ac{fcct} and \ac{gvc} corpora test systems on different, yet equally important parts of the overall task of performing CDCR in news text (cf. our requirements posed in \Cref{sec:cdevcr_system_requirements}).
Beyond established knowledge, such as \ac{ecbp} testing systems on a greater number of topics while offering low variation in event instances~\cite{vossen2018dont}, we found that:
\begin{itemize}
	\item The distribution of coreference links in each corpus varies significantly, with \ac{gvc} offering roughly the same number of within-document and within-subtopic links, whereas \ac{fcct} offers many cross-subtopic links (see \Cref{tab:dataset_properties}).
	\item Structural differences between corpora (such as the number of subtopics or mentions) can pose a problem for established \ac{cdevcr} techniques such as document preclustering and lead to performance drops (see \Cref{sec:options_document_clustering}), and expose or amplify scalability issues in systems (see \Cref{sec:exp_sota_results}).
	\item Between the corpora, the relevance of the four event components (action, participants, time, location) for resolving \acl{cdevc} varies strongly. In particular, \ac{ecbp} stands out for requiring inference over event actions almost exclusively (see \Cref{sec:exp_signals}).
\end{itemize}

This means that by designing a system against a single corpus, significant aspects of \ac{cdevcr} are disregarded. Doing so introduces a bias towards corpora with specific properties, which severely limits a system's usefulness for downstream applications on data which exhibits different properties.
Therefore, when claiming that a system is capable of resolving \acl{cdevc} in the general case, it is imperative to report its performance on \textit{multiple} \ac{cdevcr} corpora to certify its robustness to all aspects of \ac{cdevcr} annotated therein.

Related to this finding is the recent trend of \ac{ecbp} systems applying document preclustering prior to mention-level event coreference resolution which deserves special attention.
Throughout this work, we have pointed out that by preclustering documents via TF--IDF, one can reproduce the subtopics of a corpus. At test time, this yields an increase in precision for the resolution of within-document and within-subtopic links but has the downside of precluding the resolution of cross-subtopic or cross-topic links. 
This downside is negligible on \ac{ecbp} because in this corpus, within-document and within-subtopic links outnumber cross-subtopic links by a factor of 100 (see \Cref{tab:dataset_properties}). Many recent \ac{ecbp} systems apply preclustering~\cite{cremisini2020new, meged2020paraphrasing, barhom2019revisiting, lee2012joint}, yet scores drop sharply when such a system is applied out-of-the-box on a corpus with a different distribution of coreference links (see \Cref{tab:results_sota}).
The performance boost from document preclustering therefore comes at the cost of an overspecialization on the coreference link distribution in \ac{ecbp}, which we consider to be a form of overfitting.
This is again a strong point for the evaluation of \ac{cdevcr} systems on multiple corpora by which this (inadvertent) oversight in existing systems can be revealed.\footnote{An argument could be made at this point that reporting precision and recall separately for each type of coreference link (as we did in \Cref{sec:exp_in_dataset_classifier}) were as insightful as evaluating a system on a variety of corpora with different properties. This is not the case since this such analysis is only possible for the subset of \ac{cdevcr} systems which compute a degree of coreference on link-level. For representation learning approaches or for metric learning approaches with cluster-level features, link-level analysis is not possible.}

\subsection{Evaluation Recommendations}
We summarize our findings with respect to the evaluation of \ac{cdevcr} with four recommendations for future research which pave the way for more general, comparable and reliably evaluated \ac{cdevcr} systems.

\textbf{1. \ac{cdevcr} systems should be tested on more than one corpus.} The \ac{ecbp}, \ac{fcct}, and \ac{gvc} corpora each are unique with respect to their topic structure, selection of topics and distribution of event coreference links, all of which can have an impact on the performance of a \ac{cdevcr} system. Furthermore, the importance of action, participants, time, and location information varies in each corpus.
For systems seeking to solve the \ac{cdevcr} task in general (i.e. where the application scenario does not necessitate the choice of one domain-specific corpus), this prompts for joint evaluation on multiple \ac{cdevcr} corpora to reveal a system's strengths and weaknesses. Where possible, the performance for each link type (within-document, within-subtopic, etc.) should be reported.

\textbf{2. The LEA evaluation metric~\cite{moosavi2016which} should be used as an additional performance indicator for \ac{cdevcr}.} This metric was previously shown to be more discriminative than previous coreference resolution metrics such as CoNLL F1 and takes size differences between clusters into account. We showed that cluster sizes in \ac{cdevcr} corpora vary significantly and observed score deltas between CoNLL F1 and LEA F1 of up to \SI{20}{\pp} which motivates its use for \ac{cdevcr} over CoNLL F1.

\textbf{3. In addition to a blind prediction, system performance should be reported when using gold document clusters}. \Citet{cremisini2020new} request that future system development efforts should report scores with or without document clustering. We agree with this suggestion and refine it further. Given that in the research area of within-document coreference it is commonplace to report separate scores for mention identification and coreference resolution, we think distinguishing scores obtained with and without knowledge on the gold corpus structure would only make sense for \ac{cdevcr}.	
Researchers must take care to define the gold document clusters based on the transitive closure of event coreference links (see \Cref{sec:exp_sota}). Using the topics or subtopics of a corpus for this purpose produces incorrect results since cross-subtopic (or cross-topic) coreference links in the corpus are not taken into account.

\textbf{4. Event detection performance should be evaluated carefully on \ac{cdevcr} corpora.}
From the point of view of a general-purpose event mention detection system, the event mention annotations in the \ac{ecbp}, \ac{fcct} and \ac{gvc} corpora are incomplete by design (see \Cref{sec:mention_detection_argument}). Care must be taken to not unfairly penalize a systems which includes a mention detection step as it may detect valid event mentions for which no gold annotation exists. We recommend to compute mention detection performance only on those sentences which contain gold event mention annotations and to report coreference resolution performance using the gold event mention annotations as a remedy.

\section{Future Work}
\label{sec:future_work}
Having established that evaluation on multiple of the currently available corpora is necessary for a reliable performance assessment of \ac{cdevcr} systems, we consider the development of approaches which show consistent performance in such a scenario as the next short to medium term goal for this task.

A key challenge will be achieving systems which scale to collections of 10k-100k documents without precluding the resolution of cross-subtopic and cross-topic links. A foundation has already been laid by \citet{kenyon-dean2018resolving} who investigated scalable representation learning approaches for \ac{cdevcr}.
Since current corpora consist of less than 1k documents, this may require the annotation of additional corpora.
To keep the costs of annotating corpora of such magnitude manageable, novel semi-automatic annotation techniques would be required.
Furthermore, the concept of cross-topic event coreference links has not been investigated yet due to a lack of annotated data.
Once sufficient robustness and/or scalability has been achieved, use cases for downstream applications of \ac{cdevcr} could be investigated.

\section{Conclusion}
\label{sec:conclusion}

\acresetall

The usefulness of \ac{cdevcr} for downstream multi-document NLP tasks has not been demonstrated yet.
To perform well on unseen data in general, NLP systems need to robustly handle variations in the data they are applied on. For \ac{cdevcr}, multiple corpora with varying properties have been annotated, yet each \ac{cdevcr} system to date was developed, trained, and evaluated on only a single one of them. Besides hurting comparability, this currently allows little conclusions to be drawn on their robustness and generalizability, which contributes to the initially stated problem. We addressed this situation in several ways:

We eliminated the remaining hurdles which rendered joint training and evaluation on multiple \ac{cdevcr} corpora difficult by creating \acs{fcct}, a reannotation and extension of the \ac{fcc} on token level.

To identify the unique properties of each corpus for resolving event coreference in practice, we developed a mention pair \ac{cdevcr} system with a broad set of handcrafted features and applied it on the \ac{ecbp}, \ac{fcct}, and \ac{gvc} corpora.
Using this system, we found that only a subset of all components by which events are commonly defined (action, participants, time, location) are required for resolving \ac{cdevcr} in each corpus in practice. In particular, \ac{ecbp} only focuses on resolving event actions whereas \ac{gvc} and \ac{fcct} are more balanced and additionally demand interpretation of event dates and its participants.
Link-level analysis of this system revealed that mention distance (with respect to the topic-subtopic-document hierarchy of a corpus) positively correlates with difficulty in resolving \acl{cdevc} links.

In the first uniform evaluation scenario involving multiple \ac{cdevcr} systems and corpora, we compared the neural \ac{ecbp} system of \citet{barhom2019revisiting} to the feature-based system. Firstly, we found that the neural system performs well on \ac{gvc} but is outperformed by the conceptually simpler mention pair approach on \ac{fcct}.
Secondly, we deduced from these experiments that systems which are developed for \ac{ecbp} and which apply document preclustering overfit to the link distribution in this corpus.

In brief experiments with joint training on multiple corpora, we achieve a combined LEA F1 of \num{40.9} across all three corpora with the feature-based system -- over \SI{4.5}{\pp} better than the same system trained on either corpus in isolation.

We offered four recommendations for future research on \ac{cdevcr}. Most importantly, we advocate evaluation on multiple corpora after having provided conclusive evidence that evaluating on a single corpus is and was insufficient.

All in all, with our annotation effort, corpus analyses, experiments, and open source implementation we have laid a solid foundation for future research on robust and general \ac{cdevcr} systems.
Achieving such systems then constitutes a big step forward towards \ac{cdevcr} becoming an integral part of the multi-document NLP pipeline.

\begin{acknowledgments}
	We thank Mohsen Mesgar, Kevin Stowe, Prasetya Ajie Utama and Mingzhu Wu for their helpful comments. Special thanks are due to Jan-Christoph Klie and Nafise Sadat Moosavi for the frequent exchange of ideas.
	This work was supported by the German Research Foundation through the German--Israeli Project Cooperation (DIP, grant DA 1600/1--1 and grant GU 798/17--1).
\end{acknowledgments}

\clearpage

\appendix

\appendixsection{Full In-Dataset \acs{cdevcr} Results}
\label{sec:app_extended_in_dataset_results}

\begin{table}[ht]
	\centering
	\rotatebox{90}{%
		\small
		\setlength{\tabcolsep}{3.5pt}
		\begin{tabular}{@{}lll*{13}{S[table-format=2.1, detect-weight=true]}@{}}
			\toprule
			Corpus		& System    			& Preclustering	& \multicolumn{3}{c}{MUC}	& \multicolumn{3}{c}{B$^3$}	& \multicolumn{3}{c}{CEAFe}	& \multicolumn{1}{c}{CoNLL} & \multicolumn{3}{c}{LEA} \\
			\cmidrule(lr){4-6}
			\cmidrule(lr){7-9}
			\cmidrule(lr){10-12}
			\cmidrule(lr){13-13}
			\cmidrule(l){14-16}
			&						&				& {P}		& {R}		& {F1}		& {P}		& {R}		& {F1}		& {P}		& {R}		& {F1}		& {F1}		& {P}		& {R}		& {F1} \\
			\midrule
			\acs{ecbp}	& lemma					& n/a			& 59.7015	& 69.7436	& 64.333	& 58.0893	& 67.8823	& 62.6051	& 66.297	& 52.7906	& 58.7779	& 61.9053	& 42.8045	& 43.4982	& 43.1486 \\
			& lemma-$\delta$		& n/a			& 81.4994	& 69.1282	& 74.8058	& 88.4451	& 67.6869	& 76.6861	& 66.1365	& 78.2957	& 71.7043	& 74.3987	& 71.5428	& 53.6916	& 61.3449 \\
			\addlinespace
			& feature-based	& none			& 76.0833	& 76.0205	& 76.0518	& 81.1882	& 71.7779	& 76.1934	& 72.1489	& 72.2205	& 72.1845	& 74.8099	& 67.8531	& 55.0546	& 60.7871 \\
			& 						& k-means		& 77.3523	& 76.2256	& 76.7848	& 82.3109	& 71.8443	& 76.7221	& 71.963	& 73.2325	& 72.5922	& 75.3664	& 68.9013	& 55.6042	& 61.5423 \\
			& 						& gold			& 79.1648	& 76.6154	& 77.8692	& 84.6904	& 72.011	& 77.8374	& 71.6797	& 74.4756	& 73.0509	& 76.2525	& 71.5698	& 56.3787	& 63.0718 \\
			\addlinespace
			& \acs{ba2019}			& k-means		& 80.67		& 83.49		& 82.06		& 78.25		& 81.39		& 79.79		& 77.46		& 74.19		& 75.79		& 79.21		& 67.17		& 68.04		& 67.60 \\
			&						& gold			& 80.26		& 83.79		& 81.99		& 77.48		& 81.79		& 79.58		& 77.80		& 73.65		& 75.67		& 79.08		& 66.67		& 68.50		& 67.57 \\
			\midrule
			\acs{fcct}	& lemma					& n/a			& 66.1972	& 58.75		& 62.2517	& 59.2947	& 33.269	& 42.6231	& 19.433	& 31.0226	& 23.8967	& 42.9238	& 38.3587	& 19.8697	& 26.1788 \\
			& lemma-$\delta$		& n/a			& 66.1972	& 58.75		& 62.2517	& 59.2947	& 33.269	& 42.6231	& 19.433	& 31.0226	& 23.8967	& 42.9238	& 38.3587	& 19.8697	& 26.1788 \\
			& lemma-time			& n/a			& 64.4509	& 50.6818	& 56.743	& 64.6777	& 27.6343	& 38.7235	& 17.5634	& 37.4544	& 23.9132	& 39.7932	& 36.8327	& 14.2433	& 20.5427 \\
			\addlinespace
			& feature-based	& none/gold		& 78.2773	& 82.7273	& 80.4404	& 38.3255	& 70.8046	& 49.1666	& 40.3836	& 28.1848	& 33.1891	& 54.2654	& 30.4189	& 60.4073	& 39.7645 \\
			\addlinespace
			& \acs{ba2019}			& none/gold		& 84.71		& 49.09		& 62.16		& 83.04		& 36.01		& 50.23		& 20.79		& 67.12		& 31.74		& 48.05		& 39.94		& 27.18		& 32.35 \\
			& 						& k-means		& 78.33		& 34.09		& 47.51		& 88.25		& 17.94		& 29.81		& 16.57		& 66.18		& 26.50		& 34.61		& 36.93		& 08.56		& 13.90 \\
			\midrule
			\acs{gvc}	& lemma					& n/a			& 52.0759	& 57.9156	& 54.8407	& 18.4373	& 44.2145	& 26.0231	& 28.6073	& 16.0082	& 20.5288	& 33.7975	& 08.80542	& 29.7228	& 13.586 \\
			& lemma-$\delta$		& n/a			& 70.1754	& 52.7704	& 60.241	& 70.226	& 41.5697	& 52.2252	& 28.9452	& 57.1406	& 38.4256	& 50.2972	& 43.8354	& 28.6605	& 34.6598 \\
			& lemma-time			& n/a			& 82.0084	& 51.715	& 63.4304	& 85.4013	& 40.5709	& 55.0091	& 25.3173	& 62.047	& 35.9612	& 51.4669	& 53.813	& 27.2605	& 36.1887 \\
			\addlinespace
			& feature-based	& none			& 78.0916	& 66.3061	& 71.7177	& 73.5503	& 49.9092	& 59.4658	& 38.2385	& 60.9018	& 46.9794	& 59.3876	& 56.5137	& 38.1787	& 45.5702 \\
			& 						& k-means		& 83.55		& 66.2005	& 73.8701	& 81.3218	& 49.4551	& 61.5058	& 36.3916	& 66.0696	& 46.9322	& 60.7694	& 63.866	& 38.5545	& 48.0825 \\
			& 						& gold			& 85.3124	& 68.8127	& 76.1793	& 82.7209	& 51.7024	& 63.6328	& 38.186	& 67.1913	& 48.6967	& 62.8363	& 66.1115	& 41.0781	& 50.6714 \\
			\addlinespace
			& \acs{ba2019}			& k-means		& 85.23		& 89.05		& 87.10		& 66.01		& 80.95		& 72.72		& 66.38		& 54.69		& 59.97		& 73.26		& 58.68		& 74.31		& 65.58 \\
			& 						& gold			& 88.90		& 91.95		& 90.40		& 75.47		& 86.21		& 80.48		& 73.85		& 63.90		& 68.52		& 79.80		& 69.49		& 80.89		& 74.76 \\
			\bottomrule
		\end{tabular}
	}
	\caption{Full \acl{cdevcr} results for the in-dataset scenario. We report cross-document performance (all documents merged into one meta document before scoring). The scores of the feature-based system are the mean of five independent runs. Scores for \acs{ba2019} stem from a single run.}
\end{table}

\appendixsection{Full Cross-Dataset \acs{cdevcr} Results}
\label{sec:app_full_cross_dataset_results}

\begin{table}[ht]
	\centering
	\small
	\setlength{\tabcolsep}{3.25pt}
	\begin{tabular}{@{}*{5}{c@{\hspace{2pt}}}@{\hspace{2.85pt}}l*{13}{S[table-format=2.1, detect-weight=true]}@{}}
		\toprule
		\multicolumn{5}{c}{Train + Dev}						& \multicolumn{1}{c}{Test}	& \multicolumn{13}{c}{Metrics} \\
		\cmidrule(r){1-5}
		\cmidrule(r){6-6}
		\cmidrule(l){7-19}
		\multirow[b]{3}{*}{\rotatebox[origin=c]{90}{\parbox[c]{1cm}{\acs{ecbp}}}}	& \multirow[b]{3}{*}{\rotatebox[origin=c]{90}{\parbox[c]{1cm}{\acs{ecbp}$_\text{sports}$}}}	& \multirow[b]{3}{*}{\rotatebox[origin=c]{90}{\parbox[c]{1cm}{\acs{ecbp}$_\text{guns}$}}}	& \multirow[b]{3}{*}{\rotatebox[origin=c]{90}{\parbox[c]{1cm}{\acs{fcct}}}}	& \multirow[b]{3}{*}{\rotatebox[origin=c]{90}{\parbox[c]{1cm}{\acs{gvc}}}} &	&	&	&	&	&	&	&	&	&	&	&	&	& \\
		\addlinespace
		&	&	&	&	&	&	\multicolumn{3}{c}{MUC}	& \multicolumn{3}{c}{B$^3$}	& \multicolumn{3}{c}{CEAFe}	& \multicolumn{1}{c}{CoNLL} & \multicolumn{3}{c}{LEA} \\
		\cmidrule(lr){7-9}
		\cmidrule(lr){10-12}
		\cmidrule(lr){13-15}
		\cmidrule(lr){16-16}
		\cmidrule(l){17-19}
		&	&	&	&	&				& {P}		& {R}		& {F1}		& {P}		& {R}		& {F1}		& {P}		& {R}		& {F1}		& {F1}		& {P}		& {R}		& {F1} \\
		\midrule
		\multirow[c]{3}{*}{\checkmark}	& \multirow[c]{3}{*}{\plhdr}		& \multirow[c]{3}{*}{\plhdr}		& \multirow[c]{3}{*}{\plhdr}		& \multirow[c]{3}{*}{\plhdr}		& \acs{ecbp}	& 76.08334246827484	& 76.02051282051282	& 76.05176786443771	& 81.18816005100872		& 71.77791757099594		& 76.19337460407268		& 72.14890246783381		& 72.22048056640046		& 72.18446884668902		& 74.80987043839981		& 67.85305954968877		& 55.05455494947548		& 60.787136826441 \\
		&									&									&									&									& \acs{fcct}	& 62.91585550363517	& 44.84090909090909	& 52.36217035788266	& 71.00965811860209		& 21.638112692379868	& 33.16828996030143		& 18.817967399405733	& 47.47766458016652		& 26.95273152491029		& 37.494397281031466	& 41.42023397622781		& 08.524460588440183	& 14.134440954440702 \\
		&									&									&									&									& \acs{gvc}		& 55.37144152324271	& 51.68865435356201	& 53.46638696071906	& 50.27606733054479		& 40.09163903925749		& 44.60857308515195		& 33.31116954451189		& 42.01392549402305		& 37.15796379698493		& 45.07764128095197		& 32.096465241209404	& 25.26161301075749		& 28.27132251891441 \\
		\addlinespace
		\multirow[c]{3}{*}{\plhdr}		& \multirow[c]{3}{*}{\plhdr}		& \multirow[c]{3}{*}{\plhdr}		& \multirow[c]{3}{*}{\checkmark}	& \multirow[c]{3}{*}{\plhdr}		& \acs{ecbp}	& 57.31447925247659	& 91.13846153846156	& 70.37176869811134	& 22.102907883145884	& 88.96737695466175		& 35.367857846294576	& 72.23175621362551		& 20.602364180457056	& 32.03864441049266		& 45.92609031829954		& 15.96131731569325		& 57.15711386994929		& 24.907385103300675 \\
		&									&									&									&									& \acs{fcct}	& 78.27726422124033	& 82.72727272727272	& 80.44039054722198	& 38.32545963848727		& 70.80459210505688		& 49.16662911373729		& 40.38356224096661		& 28.18483948704755		& 33.189115364870025	& 54.26537834194309		& 30.41894024174348		& 60.40733495568591		& 39.76453832193483 \\
		&									&									&									&									& \acs{gvc}		& 73.46318042018662	& 88.97097625329815	& 80.47673771687016	& 06.443801507038814	& 82.90827672152593		& 11.95586061316573		& 31.00154457279721		& 05.304055122736693	& 09.055373022723448	& 33.82932378425311		& 03.5526492442660144	& 74.9774494149505		& 06.781433642264527 \\
		\addlinespace
		\multirow[c]{3}{*}{\plhdr}		& \multirow[c]{3}{*}{\plhdr}		& \multirow[c]{3}{*}{\plhdr}		& \multirow[c]{3}{*}{\plhdr}		& \multirow[c]{3}{*}{\checkmark}	& \acs{ecbp}	& 64.53380891561182	& 82.52307692307692	& 72.42760389783012	& 63.51688047459509		& 78.86796960036585		& 70.36301588912485		& 68.7727013573079		& 45.55354374875151		& 54.80309457790608		& 65.86457145495368		& 48.84504909141456		& 50.16833842982555		& 49.49658000370589 \\
		&									&									&									&									& \acs{fcct}	& 65.17146483759146	& 56.65909090909091	& 60.61391217587111	& 62.58467984552054		& 30.96426009163177		& 41.3914853816344		& 22.294419100259147	& 37.723725131897534	& 28.012788685304313	& 43.33939541426995		& 42.16430166186827		& 17.65703260181437		& 24.8543418799426 \\
		&									&									&									&									& \acs{gvc}		& 78.09155967556907	& 66.30606860158312	& 71.71771014121842	& 73.55033194955158		& 49.90924361906383		& 59.46582633500889		& 38.23851256573724		& 60.90178371994102		& 46.97935441251409		& 59.38763029624712		& 56.51366580128873		& 38.17872866957683		& 45.57023651600609 \\
		\midrule
		\multirow[c]{3}{*}{\plhdr}		& \multirow[c]{3}{*}{\checkmark}	& \multirow[c]{3}{*}{\plhdr}		& \multirow[c]{3}{*}{\checkmark}	& \multirow[c]{3}{*}{\plhdr}		& \acs{ecbp}	& 67.02237846478354	& 81.66153846153847	& 73.61850831509693	& 64.2097884420755		& 77.64233330616243		& 70.27749248896189		& 73.61938008985762		& 54.13504625929141		& 62.38210697925128		& 68.75936926110338		& 52.06447308722967		& 54.27976093962945		& 53.13829704160747 \\
		&									&									&									&									& \acs{fcct}	& 75.70282217594307	& 74.56818181818182	& 75.13083677736775	& 54.73389492710325		& 51.49598749802007		& 52.95533961659649		& 33.68501629719953		& 36.355783462740543	& 34.96538157005017		& 54.35051932133814		& 42.71918700291826		& 40.24933816165808		& 41.31191612305315 \\
		&									&									&									&									& \acs{gvc}		& 52.82007112577601	& 55.80474934036939	& 54.27093796557969	& 28.6568366929974		& 41.36658723776928		& 33.84270690582593		& 34.58648852882356		& 26.912172991869643	& 30.26312921906212		& 39.45892469682258		& 16.892331067148966	& 26.61796912156743		& 20.64424301755212 \\
		\addlinespace
		\multirow[c]{3}{*}{\plhdr}		& \multirow[c]{3}{*}{\plhdr}		& \multirow[c]{3}{*}{\checkmark}	& \multirow[c]{3}{*}{\plhdr}		& \multirow[c]{3}{*}{\checkmark}	& \acs{ecbp}	& 72.26073266759354	& 81.00512820512822	& 76.38327035426354	& 76.60975367622759		& 77.25717320291583		& 76.93164376455226		& 71.19016190069071		& 60.75460424554828		& 65.55925109756394		& 72.95805507212658		& 62.54681323670088		& 55.16719422923655		& 58.62544027960724 \\
		&									&									&									&									& \acs{fcct}	& 64.98917805305408	& 59.27272727272728	& 61.99865276716514	& 59.30881795446077		& 33.39079480921215		& 42.72436874698139		& 23.32964967531779		& 34.21048308412544		& 27.737551067836125	& 44.15352419399422		& 41.478215050120704	& 19.94434569437878		& 26.93465376528754 \\
		&									&									&									&									& \acs{gvc}		& 69.50015292505116	& 63.61477572559366	& 66.42710579763762	& 60.90155202630705		& 48.06204310126726		& 53.72476572006348		& 39.34888477039368		& 52.43544598175465		& 44.95799101896151		& 55.03662084555421		& 45.016429570795546	& 35.44894706146014		& 39.66339778359211 \\
		\addlinespace
		\multirow[c]{3}{*}{\checkmark}	& \multirow[c]{3}{*}{\plhdr}		& \multirow[c]{3}{*}{\plhdr}		& \multirow[c]{3}{*}{\checkmark}	& \multirow[c]{3}{*}{\plhdr}		& \acs{ecbp}	& 73.20324011791646	& 75.63076923076923	& 74.39588689034597	& 77.23967788116429		& 71.83840021237804		& 74.43983237672848		& 72.22195484490774		& 69.31465094973228		& 70.73629324563044		& 73.19067083756831		& 63.25672317658109		& 53.60701338697687		& 58.03288627726739 \\
		&									&									&									&									& \acs{fcct}	& 71.34074578845692	& 58.72727272727272	& 64.42098738880201	& 71.57815581389559		& 31.0029119422468		& 43.261369062324906	& 27.952190219921924	& 54.14985049526475		& 36.86758018719268		& 48.18331221277321		& 50.92106676240623		& 19.69907844434076		& 28.40139622328898 \\
		&									&									&									&									& \acs{gvc}		& 56.95746796488143	& 54.37994722955144	& 55.638006180158	& 46.47424096027402		& 41.18192188042859		& 43.65896453404502		& 35.05884026332318		& 41.26830830570448		& 37.90524082114049		& 45.73407051178117		& 29.65042353403821		& 26.674242869041537	& 28.07149001909567 \\
		\addlinespace
		\multirow[c]{3}{*}{\checkmark}	& \multirow[c]{3}{*}{\plhdr}		& \multirow[c]{3}{*}{\plhdr}		& \multirow[c]{3}{*}{\plhdr}		& \multirow[c]{3}{*}{\checkmark}	& \acs{ecbp}	& 68.8161535334		& 81.88717948717947	& 74.78450770827231	& 68.52157197913422		& 78.0636432426311		& 72.98111594808427		& 73.08428810428061		& 56.26949162649929		& 63.58308720099651		& 70.44957028578437		& 55.37047479809182		& 54.77442546022191		& 55.07036420021515 \\
		&									&									&									&									& \acs{fcct}	& 65.37932530847638	& 66.13636363636364	& 65.75540467291956	& 50.0217965462197		& 39.20456046860036		& 43.95251714573768		& 25.90478375221224		& 24.315127604172815	& 25.08163462629094		& 44.92985214831605		& 38.00609962253095		& 25.46761266709582		& 30.49346826332583 \\
		&									&									&									&									& \acs{gvc}		& 60.35266563123888	& 61.76781002638523	& 61.05185216720294	& 39.99923418423868		& 46.36150858806755		& 42.93797971881693		& 38.33750161012997		& 34.809264709744203	& 36.486104408936604	& 46.82531209831883		& 27.359811431603875	& 32.83516539986092		& 29.84021388543425 \\
		\addlinespace
		\multirow[c]{3}{*}{\plhdr}		& \multirow[c]{3}{*}{\plhdr}		& \multirow[c]{3}{*}{\plhdr}		& \multirow[c]{3}{*}{\checkmark}	& \multirow[c]{3}{*}{\checkmark}	& \acs{ecbp}	& 59.8041671675956	& 81.74358974358974	& 69.0712511415144	& 50.62399521758748		& 78.15016631649445		& 61.43206940741839		& 68.43330001499914		& 38.01865670977261		& 48.86681869254229		& 59.79004641382504		& 37.56206667064297		& 47.74465854529305		& 42.03851253552827 \\
		&									&									&									&									& \acs{fcct}	& 70.27681909537767	& 78.02272727272727	& 73.9472899489923	& 39.59010979129494		& 61.01936423630657		& 48.0180954842694		& 38.60322880959189		& 16.045507124865893	& 22.658834601966965	& 48.20807334507623		& 34.10112696249451		& 48.51992369849193		& 40.04400288717195 \\
		&									&									&									&									& \acs{gvc}		& 71.05653359196805	& 65.09234828496042	& 67.94292948538851	& 60.66993990177231		& 48.79703563043506		& 54.08662476251658		& 39.51978736306397		& 52.54132669252682		& 45.10542034601445		& 55.71165819797318		& 45.69485388518007		& 36.55810043515443		& 40.61763923438674 \\
		\addlinespace
		\multirow[c]{3}{*}{\checkmark}	& \multirow[c]{3}{*}{\plhdr}		& \multirow[c]{3}{*}{\plhdr}		& \multirow[c]{3}{*}{\checkmark}	& \multirow[c]{3}{*}{\checkmark}	& \acs{ecbp}	& 56.80427400403149	& 89.98974358974359	& 69.64550865175967	& 32.25403346482604		& 87.1657853522344		& 47.07355613389462		& 65.78452835951423		& 19.240265846576135	& 29.76718991903372		& 48.82875156822933		& 24.28429177381188		& 53.595104615098		& 33.411812635837485 \\
		&									&									&									&									& \acs{fcct}	& 73.57970700585682	& 84.43181818181819	& 78.63301766141421	& 29.59756058771296		& 70.68960342957464		& 41.72093487457661		& 35.303879317654474	& 07.706267763276911	& 12.644387441706034	& 44.33277999256561		& 26.49204062441818		& 58.80177695888781		& 36.52164036386481 \\
		&									&									&									&									& \acs{gvc}		& 63.9228778588282	& 66.80738786279684	& 65.33325416796532	& 42.64786906803544		& 50.78118747636381		& 46.35849808488965		& 43.03266069686306		& 35.40552636282487		& 38.84744950872556		& 50.17973392052684		& 32.313692176180625	& 38.26419839500644		& 35.03191382326289 \\
		\bottomrule
	\end{tabular}
	\caption{Full \acl{cdevcr} results for the cross-dataset scenario. We report cross-document performance (all documents merged into one meta document before scoring). The presented scores are the mean of five independent runs.}
\end{table}

\clearpage

\appendixsection{Mention Pair Features}
\label{sec:app_feature_details}

\newlist{features}{itemize}{1}
\setlist[features]{label=\textbullet, nosep, leftmargin=9pt, before=\vspace{-\baselineskip}\ttfamily}

{%
	\centering
	\small
	\begin{longtable}{@{}p{1.9cm}p{4.7cm}p{5.9cm}@{}}
		\toprule
		Type	& Features	& Description \\
		\midrule
		\endhead
		action mention string distance	& \begin{features}
			\item \seqsplit{is-surface-form-identical}
			\item \seqsplit{is-lemma-identical}
			\item \seqsplit{surface-form-mlipns-distance}
			\item \seqsplit{surface-form-levenshtein-distance}
		\end{features}	&  String distance on two action mentions. We compare surface form and lemma for identity and compute Levenshtein and MLIPNS~\cite{shannaq2010using} distances for lexical and phonetic distances. \\
		\midrule
		TF--IDF							& \begin{features}
			\item \seqsplit{document-similarity}
			\item \seqsplit{surrounding-sentence-similarity}
			\item \seqsplit{sentence-context-similarity}
		\end{features} & We fit TF--IDF vectors on all given documents. We then compute the cosine similarity between TF-IDF vectors of text regions belonging to two mention pairs. For the regions we use (1) the full document, (2) the sentence surrounding the event mention, and (3) a sentence window of 5 sentences surrounding the mention (i.e., its context). \\
		\midrule
		sentence embedding similarity	& \begin{features}
			\item \seqsplit{surrounding-sentence}
			\item \seqsplit{doc-start}
		\end{features} & We compute the cosine similarity between sentence representations of a sentence pair originating from a mention pair. We compare the sentences surrounding each event mention and the first sentence of a mention's document. Sentence representions are computed with the Sentence-BERT framework~\cite{reimers2019sentence}, using the pretrained \texttt{distilbert\hyp{}base\hyp{}nli\hyp{}stsb\hyp{}mean\hyp{}tokens} model. \\
		\midrule
		action mention embedding similarity   & \begin{features}
			\item \seqsplit{action-mention}
		\end{features}			& We compute a contextualized span representation of the action mention of each event mention and compute the cosine similarity of these representations for each mention pair. Span representations are created from the pretrained SpanBERT large model~\cite{joshi2020spanbert} using a window of five sentences surrounding each event mention. \\
		\midrule
		spatial distance				& \begin{features}
			\item \seqsplit{distance-document-level-\{geo-hierarchy-match/geodesic-distance\}}
			\item \seqsplit{distance-srl-level-\{geo-hierarchy-match/geodesic-distance\}}
			\item \seqsplit{distance-sentence-level-\{geo-hierarchy-match/geodesic-distance\}}
			\item \seqsplit{distance-closest-preceding-sentence-level-\{geo-hierarchy-match/geodesic-distance\}}
			\item \seqsplit{distance-closest-overall-level-\{geo-hierarchy-match/geodesic-distance\}}
		\end{features} & We obtain a location from each mention in five ways: (1) document-level, where we pick the first entity-linked location in a document, (2) \acs{srl}-level, where we use \acf{srl} to find the linked location expression attached to the mention action, (3) sentence-level, where we use the location expression closest to the mention action in the same sentence, (4) closest-sentence-level, where we use the closest preceding location expression from all previous sentences and (5), a combination which applies (2), (3), (4) in order until a location expression is found.
		
		For each location pair, we compute distances in two ways: (1) We compute the geodesic distance between the coordinates of both locations. (2) For each location, we follow the \texttt{subdivision} and \texttt{country} relations in DBpedia upwards (from more specific to less specific locations) to find a match between the two locations. The earlier a match is found, the smaller is the distance between the two locations. \\
		\midrule
		temporal distance				& \begin{features}
			\item \seqsplit{distance-document-publish-level-\{year/month/week/day/hour\}}
			\item \seqsplit{distance-document-level-\{year/month/week/day/hour\}}
			\item \seqsplit{distance-srl-level-\{year/month/week/day/hour\}}
			\item \seqsplit{distance-sentence-level-\{year/month/week/day/hour\}}
			\item \seqsplit{distance-closest-preceding-sentence-level-\{year/month/week/day/hour\}}
			\item \seqsplit{distance-closest-overall-level-\{year/month/week/day/hour\}}
		\end{features} & Computes temporal distance between temporal expressions belonging to a mention pair on different date fields (difference ofdays, difference of hours, \ldots). Multiple variants for finding temporal expressions exist: (1) the document publication date (where available), (2) document-level, where we pick the first temporal expression in a document, (3) \acs{srl}-level, where we use \ac{srl} to find the temporal expression attached to the mention action, (4) sentence-level, where we use the temporal expression closest to the mention action in the same sentence, (5) closest-sentence-level, where we use the closest preceding temporal expression from all previous sentences and (6), a combination which applies (3), (4), (5) in order until a temporal expression is found. \\
		\midrule
		Wikidata embedding similarity	& \begin{features}
			\item \seqsplit{action-mention}
			\item \seqsplit{semantic-role-args-\{mean/variance/min/max\}}
			\item \seqsplit{surrounding-sentence-\{mean/variance/min/max\}}
			\item \seqsplit{sentence-context-\{mean/variance/min/max\}}
			\item \seqsplit{doc-start-\{mean/variance/min/max\}}
		\end{features} & We obtain Wikidata QIDs for each DBpedia entity and map these to pretrained embeddings from PyTorch-BigGraph~\cite{lerer2019pytorch}.
		For each mention in a mention pair, we look up the vectors of (1) the linked action mention (where available), (2) of all event components, (3) of all linked entities in the surrounding sentence, (4) of all linked entities in a 5-sentence window around the mention and (5) of all linked entities in the first three document sentences. Between each of these groups, we compute the pairwise cosine similarity between all vectors and retain the mean, variance, minimum and maximum similarity respectively. \\
		\bottomrule
	\end{longtable}
	\captionof{table}{Feature Overview}
	\label{tab:feature_overview}
	\par
}

\appendixsection{Corpus Splits}
\label{sec:app_corpus_splits}
\begin{description}
	\item[\ac{ecbp}:] We used the official splits defined in the corpus readme file and filtered the sentences according to the extra CSV file provided with the corpus. In \Cref{sec:exp_sota} we only compare systems also using these splits.
	\item[\ac{fcct}:] We used the tournaments 2010, 2012 and 2014 for training, 2016 for development and 2018 for testing. We remove all mentions belonging to the \texttt{other\_event} cluster (see \Cref{sec:annotation_results}).
	\item[\ac{gvc}:] No official splits are provided for this corpus. We compiled a list of all 241 gun violence incidents present in the corpus and the mapping from incident to document. We shuffled the incidents and partitioned them randomly so that train, dev and test contain 70/15/15 percent of all documents respectively. The list of documents in each split is provided alongside our system implementation. We remove all mentions with cluster ID 0: this cluster contains mentions of generic events or unresolved cross-subtopic coreference links.
\end{description}

\appendixsection{Feature Importance}
\label{sec:app_feature_importance}

{%
	\centering
	\small
	\setlength{\tabcolsep}{3pt}
	\begin{longtable}{@{}lS[table-format=0.3]>{\ttfamily}ll@{}}
		\toprule
		Corpus		& {Gain}		& \rmfamily Feature 									& Type \\
		\midrule
		\endhead
		\acs{ecbp}	& 0.342106		& is-lemma-identical									& action string distance \\
		& 0.248146		& surface-form-mlipns-distance							& action string distance \\
		& 0.130495		& action-mention										& action mention embedding \\
		& 0.0850028		& is-surface-form-identical								& action string distance \\
		& 0.0555572		& document-similarity									& TF--IDF \\
		& 0.0517218		& context-similarity									& TF--IDF \\
		& 0.0427347		& surface-form-levenshtein-distance						& action string distance \\
		& 0.0276008		& surrounding-sentence-similarity						& TF--IDF \\
		& 0.0166361		& doc-start												& sentence embedding \\
		\midrule
		\acs{fcct}	& 0.301784		& surface-form-mlipns-distance							& action string distance \\
		& 0.144558		& is-lemma-identical									& action string distance \\
		& 0.0983997		& distance-closest-overall-level-year					& temporal distance \\
		& 0.0805005		& is-surface-form-identical								& action string distance \\
		& 0.0534523		& distance-sentence-level-year							& temporal distance \\
		& 0.0524243		& surface-form-levenshtein-distance						& action string distance \\
		& 0.0515036		& action-mention										& action mention embedding \\
		& 0.0412502		& surrounding-sentence-similarity						& TF--IDF \\
		& 0.0315895		& semantic-role-args-min								& Wikidata embedding \\
		& 0.0247313		& distance-closest-overall-level-week					& temporal distance \\
		& 0.0241453		& semantic-role-args-variance							& Wikidata embedding \\
		& 0.0229691		& surrounding-sentence-min								& Wikidata embedding \\
		& 0.0197022		& semantic-role-args-mean								& Wikidata embedding \\
		& 0.0193804		& sentence-context-variance								& Wikidata embedding \\
		& 0.0177014		& surrounding-sentence-variance							& Wikidata embedding \\
		& 0.0159086		& surrounding-sentence-mean								& Wikidata embedding \\
		\midrule
		\acs{gvc}	& 0.254793		& surface-form-mlipns-distance							& action string distance \\
		& 0.126396		& distance-document-publish-level-week					& temporal distance \\
		& 0.125259		& distance-document-publish-level-month					& temporal distance \\
		& 0.075458		& document-similarity									& TF--IDF \\
		& 0.0593222		& is-lemma-identical									& action string distance \\
		& 0.0409444		& is-surface-form-identical								& action string distance \\
		& 0.0355292		& distance-document-level-week							& temporal distance \\
		& 0.0345213		& action-mention										& action mention embedding \\
		& 0.0282571		& distance-document-level-month							& temporal distance \\
		& 0.0232628		& surface-form-levenshtein-distance						& action string distance \\
		& 0.0200447		& doc-start-min											& Wikidata embedding \\
		& 0.013317		& context-similarity									& TF--IDF \\
		& 0.0120535		& distance-closest-overall-level-week					& temporal distance \\
		& 0.0105536		& doc-start												& sentence embedding \\
		& 0.00934956	& distance-closest-overall-level-year					& temporal distance \\
		& 0.00918212	& distance-closest-pr-sent-level-week					& temporal distance \\
		& 0.00862737	& distance-closest-overall-level-month					& temporal distance \\
		& 0.00861989	& distance-closest-pr-sent-level-year					& temporal distance \\
		& 0.00828474	& distance-closest-pr-sent-level-month					& temporal distance \\
		& 0.00825124	& doc-start-variance									& Wikidata embedding \\
		& 0.00749147	& distance-sentence-level-week							& temporal distance \\
		& 0.00713295	& distance-document-publish-level-day					& temporal distance \\
		& 0.00698602	& doc-start-mean										& Wikidata embedding \\
		& 0.00692019	& surrounding-sentence-similarity						& TF--IDF \\
		& 0.00674616	& doc-start-max											& Wikidata embedding \\
		& 0.00642145	& sentence-context-max									& Wikidata embedding \\
		& 0.00620161	& sentence-context-min									& Wikidata embedding \\
		& 0.00615617	& distance-closest-pr-sent-level-day					& temporal distance \\
		& 0.00588971	& distance-document-level-day							& temporal distance \\
		& 0.00579038	& closest-pr-sent-lvl-geo-hier-match					& location \\
		& 0.00574741	& surrounding-sentence-min								& Wikidata embedding \\
		& 0.00566589	& surrounding-sentence									& sentence embedding \\
		& 0.00555567	& sentence-context-mean									& Wikidata embedding \\
		& 0.0052688		& sentence-context-variance								& Wikidata embedding \\
		\bottomrule
	\end{longtable}
	\captionof{table}{Full list of selected features and feature importance for each corpus.}
	\par
}

\appendixsection{ECB+ Publication Date Annotation}
\label{sec:app_annotation_ecbp}

The document publication date is an important piece of information for grounding temporal expressions, particularly in news text. \Ac{ecbp} is the only \ac{cdevcr} corpus covered in this work for which document publication dates were not annotated.
Source URLs of the corpus articles, from which the publication date could have been extracted automatically, are unfortunately only provided for documents in the \ac{ecbp} subtopics added at a later point by \citet{cybulska2014using}. For the initial \ac{ecb}, no URLs are present.
We manually annotated the missing date by inspecting the first few sentences of each document. \Cref{tab:ecbp_publish_date_annotation} displays the number of documents for which the publication date could be identified.

\begin{table}[ht]
	\centering
	\begin{tabular}{@{}lrr@{}}
		\toprule
		& \acs{ecb}	& \acs{ecbp} \\
		\midrule
		Date found				& 0		& 460 \\
		No date found			& 480	& 42 \\
		\bottomrule
	\end{tabular}
	\caption{Share of documents for which we identified and annotated the publication date. \acs{ecbp} refers to the set of documents added by \citet{cybulska2014using} for \acs{ecbp}.}
	\label{tab:ecbp_publish_date_annotation}
\end{table}

Unfortunately, no publication dates were found in the \ac{ecb} half of the corpus. In light of these results, we decided against using these annotations in our experiments since it may have given systems an unfair advantage in deciding whether a document belongs to one of the \ac{ecb} or \ac{ecbp} subtopics.
We nevertheless release our annotations in the hopes that they will be useful for future research.

\appendixsection{Hyperparameter Optimization Procedure}
\label{sec:app_hyperopt}
We describe our approach for optimizing the hyperparameters of the feature-based system.

We apply repeated k-fold cross-validation to obtain reliable results.
To define folds, we first partition the documents based on their topic~(\acs{ecbp}) or subtopic~(\acs{fcct}, \acs{gvc}). From these partitioned document sets, folds are created, based on which we generate mention pairs.
Compared to the naïve approach of creating folds from all possible mention pairs of a corpus split, this approach guarantees that \textit{each mention} in the respective test fold is unseen, opposed to just the mention \textit{pair} being unseen (with the two constituent mentions likely having been seen at training time), which provides a more faithful testing scenario.
By using topics or subtopics for partitioning, we ensure a high number of coreferring pairs in the folds and guide the hyperparameter search towards models which should generalize better across topics or subtopics.

\begin{figure}[ht]
	\centering
	\includegraphics{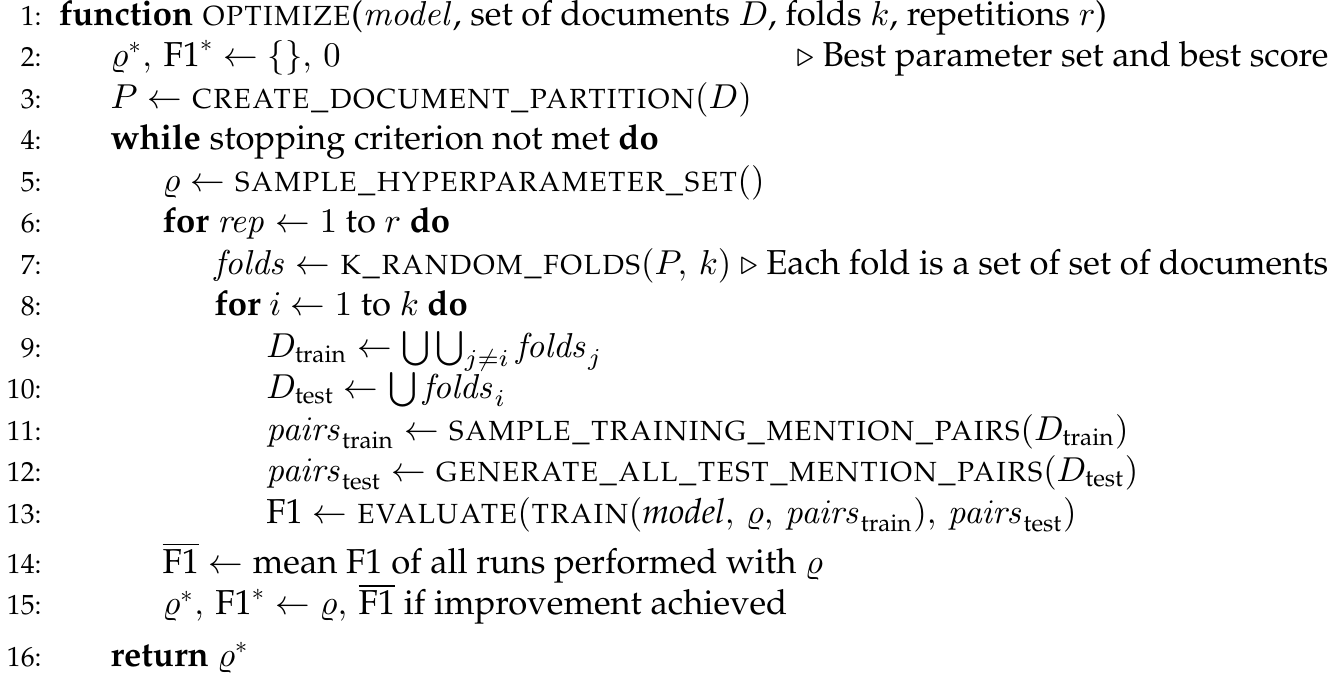}
	\caption{Hyperparameter optimization approach for the mention pair classifier}
	\label{fig:hyperopt}
\end{figure}

The optimization algorithm is shown in \Cref{fig:hyperopt}.
We use the optuna framework~\cite{akiba2019optuna} for sampling increasingly optimal sets of hyperparameters and use a configurable maximum duration as the stopping criterion.
Optimization of the agglomerative clustering step is performed similarly, with the difference of generating a test clustering in line 12 and using the LEA F1 metric instead of F1 for binary classification in line 13.

\appendixsection{Mention Pair Generation at Training Time}
\label{sec:app_mention_pair_generation}

\paragraph{Approach}
We explain our approach for determining the number of coreferring mention pairs to randomly sample for each event during training.
Given a set of events $\mathbb{E} = \{e_1, e_2, \ldots\}$ and the function $\mathrm{m}: \mathbb{E} \rightarrow \mathbb{N}$ providing the number of mentions for an event (i.e. the cluster size), we define $\mathrm{pairs}_\text{coref}: \mathbb{E} \rightarrow \mathbb{N}$, the number of coreferring pairs to sample for an event, as:
\begin{align}
	\mathrm{pairs}_\text{coref}(e) &= \left\lceil\left(\mathrm{m}(e)-1\right) \cdot \min\left(\mathrm{undersample}(e), \frac{\mathrm{m}(e)}{2}\right)\right\rceil \\
	\mathrm{undersample}(e) &= c + \mathrm{m}(e)^{1-\mathrm{cdf}\left(\mathrm{m}(e)\right)} - 1 \\
	\intertext{where $c \in \mathbb{R}$ is a hyperparameter and $\mathrm{cdf}: \mathbb{N} \rightarrow \mathbb{Q}$ is the percentage of all mentions in $\mathbb{E}$ originating from clusters up to a given size $i$:}
	\mathrm{cdf}(i) &= \frac{\sum_{\left\{e \in \mathbb{E}\mid\mathrm{m}(e) \leq i\right\}}\mathrm{m}(e)}{\sum_{e \in \mathbb{E}} \mathrm{m}(e)}
\end{align}
For the event $e_\text{largest}$ with the most mentions in the given data split, this results in $\mathrm{cdf}(\mathrm{m}(e_\text{largest})) = 1$, therefore $\mathrm{pairs}_\text{coref}(e_\text{largest}) = \left\lceil\left(\mathrm{m}(e)-1\right) \cdot c\right\rceil$. Hence, $c$ controls the amount of coreferring mention pairs sampled from large clusters. The number of pairs to sample transitions smoothly from linear to quadratic the smaller a cluster is with respect to the overall distribution of cluster sizes in the dataset.

Having sampled coreferring pairs for each event, we determine their coreference link type (within-document, within-subtopic, etc.). For each type of coreference link (within-document, within-subtopic, etc.), the maximum number of non-coreferring pairs generated is $k$ times the number of generated coreferring pairs. We ensure that the number of non-coreferring pairs increases per link type (so that within-document $<$ within-subtopic $< \ldots$).

\paragraph{Impact of Hyperparameters}
To evaluate how hyperparameters $c$ and $k$ affect performance, we perform exhaustive grid search with different choices of $c,\,k$ when training the mention pair classifier component of our feature-based approach on the \ac{ecbp} training split and evaluating it on the \ac{ecbp} development split.
We test $c \in \{2^{-3}, 2^{-2}, \ldots, 2^5\}$ and $k \in \{1, 2, 4, 8, 12, 16, 24, 32, 48\}$. The entire training split can produce $7.2 \cdot 10^6$ pairs. With the smallest choice of $(c=2^{-3}, k=1)$, $0.28 \cdot 10^6$ pairs are generated (just $4 \cdot 10^3$ pairs when excluding non-coreferring cross-topic pairs), whereas the largest choice $(c=2^5, k=48)$ yields $0.47 \cdot 10^6$ pairs ($0.2 \cdot 10^6$ pairs when excluding non-coreferring cross-topic pairs).

\begin{figure}[ht]
	\centering
	\includegraphics[width=.32\textwidth]{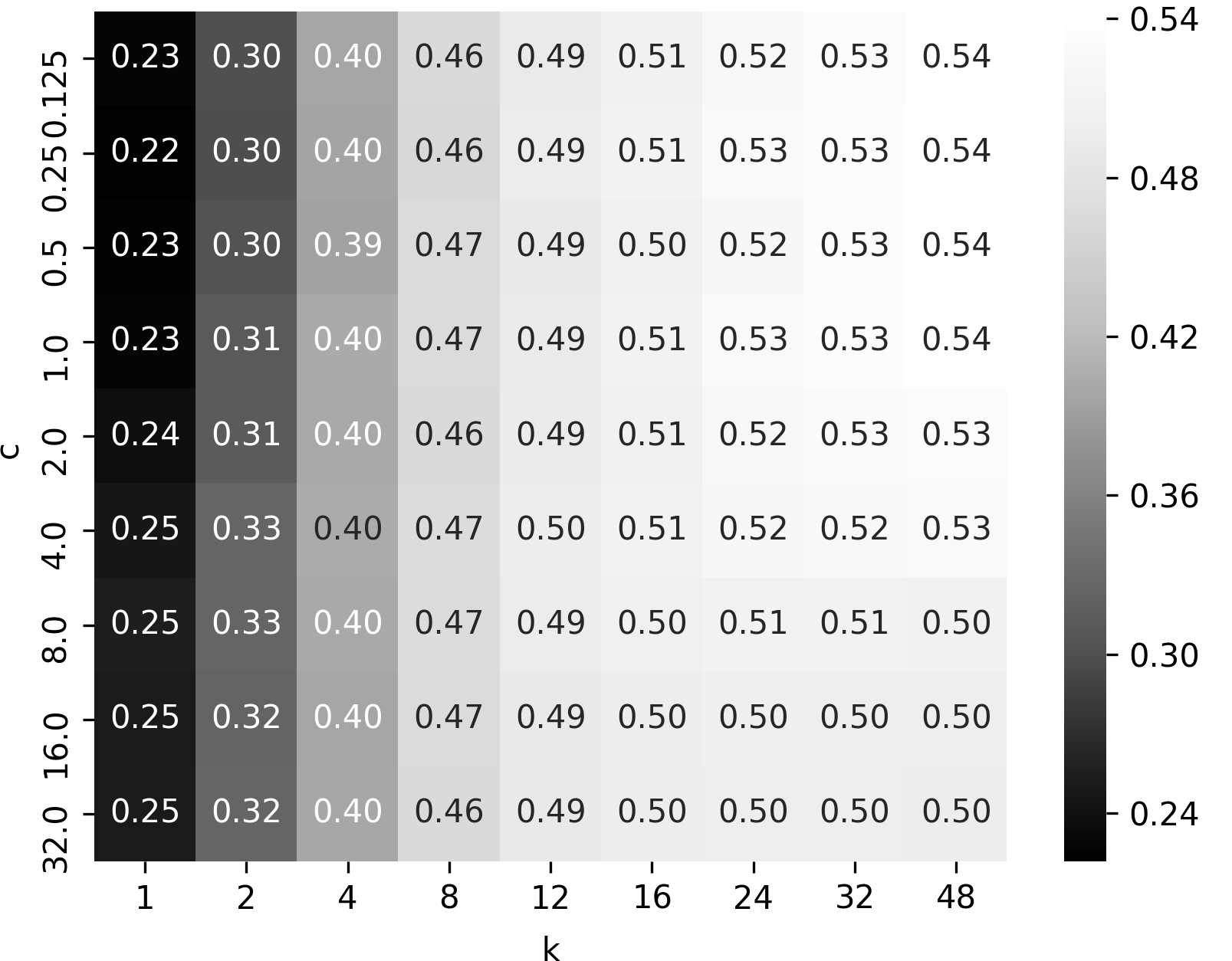}
	\hfill
	\includegraphics[width=.32\textwidth]{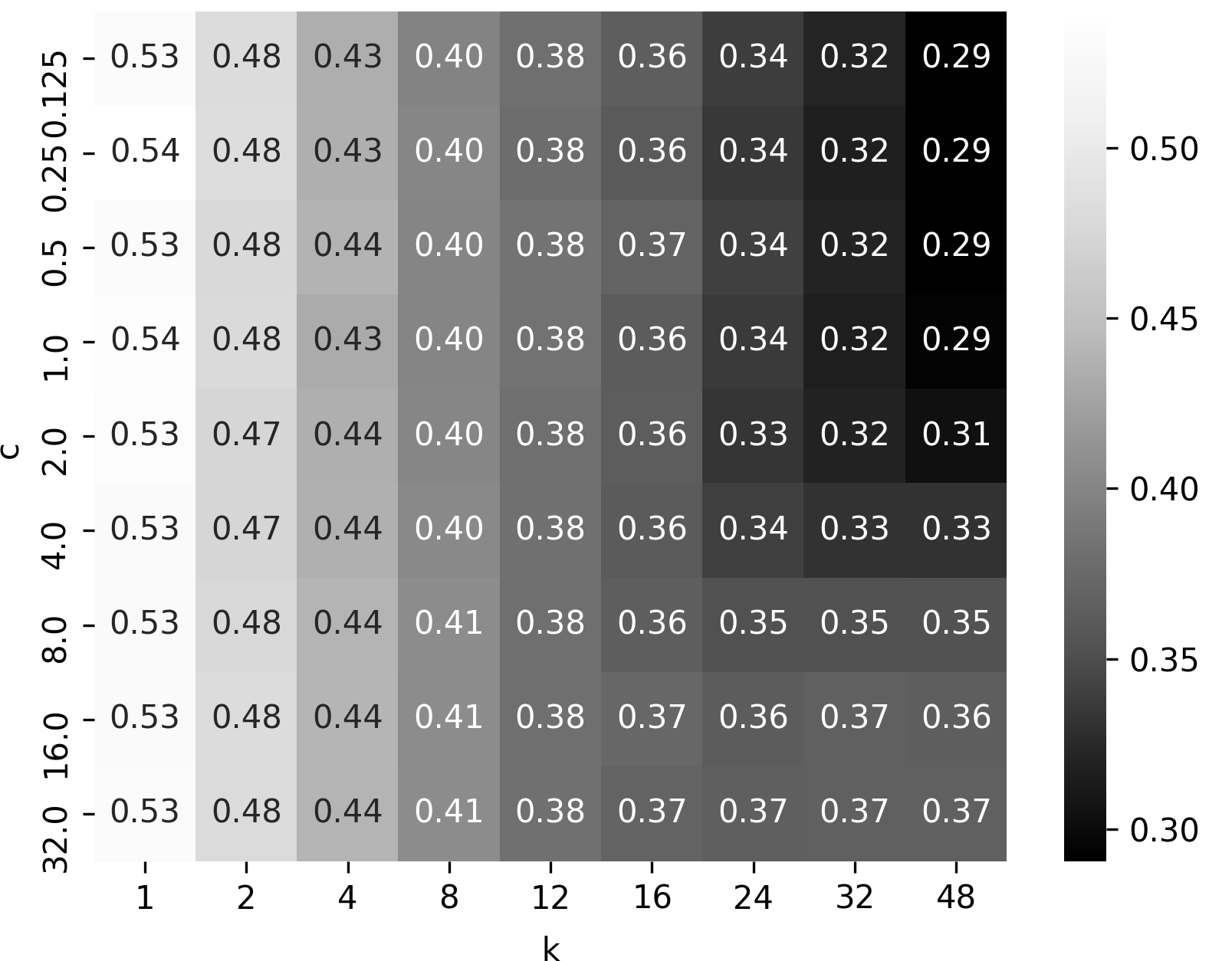}
	\hfill
	\includegraphics[width=.32\textwidth]{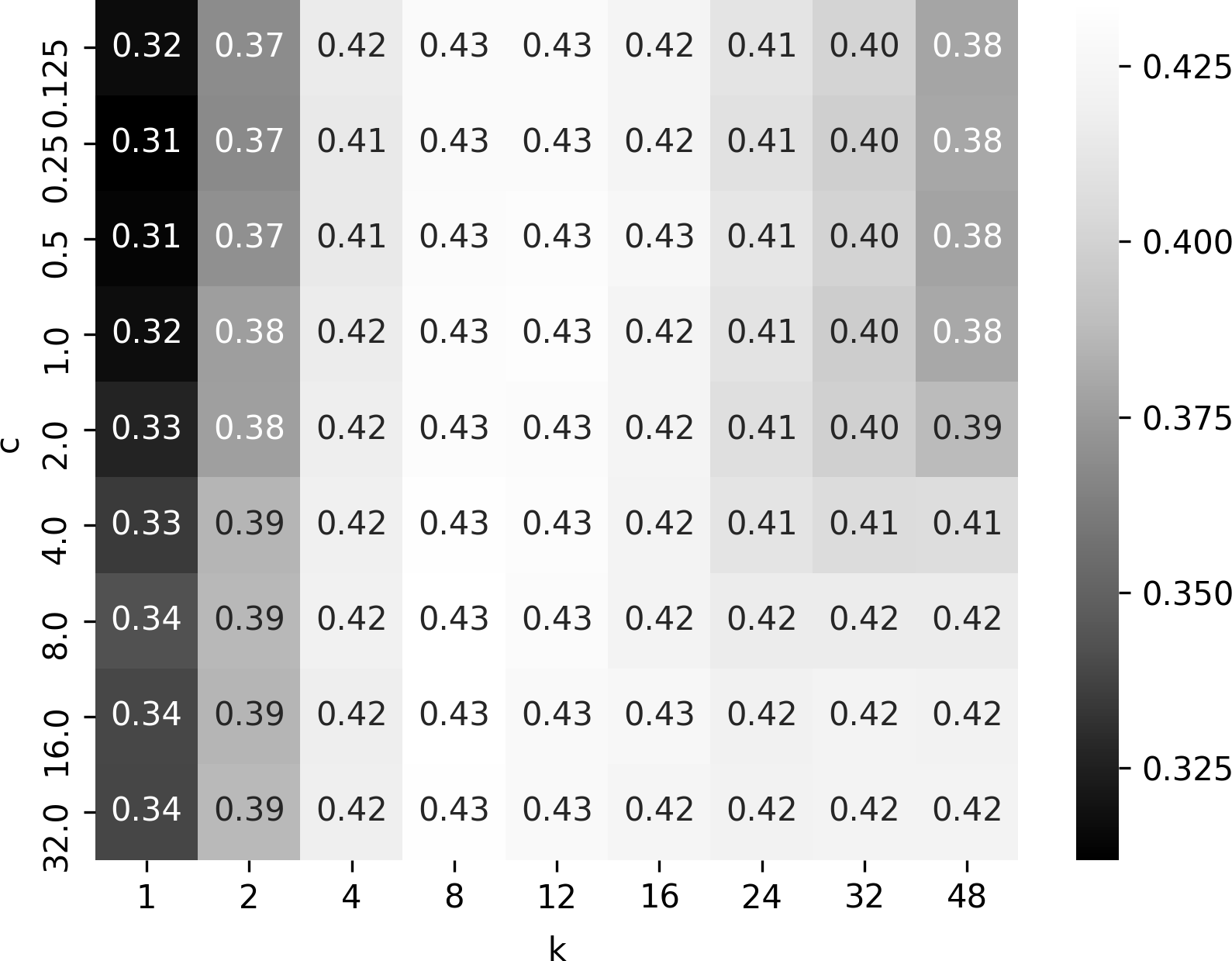}
	\caption{Mention pair classifier performance on the \ac{ecbp} development split for different choices of $c$ and $k$. From left to right: Precision, Recall, F1. "Coreferring" is used as the positive class.}
	\label{fig:undersampling_hyperparameters}
\end{figure}

For each $(c, k)$ combination, we compute precision, recall and F1 for each type of coreference link (using the mean of five independent trials to mitigate noise). We further aggregate these results by computing the macro-average over the four link types to produce one precision/recall/F1 value each per $(c, k)$ tuple, shown in \Cref{fig:undersampling_hyperparameters}.

As visualized by the plots, $k$ controls the precision/recall tradeoff, with higher $k$ (larger proportion of non-coreferring training pairs) leading to high precision but low recall.
The choice of $c$ has a smaller impact on performance. Overall, considering F1 scores, the amount of coreferring mention pairs generated from large clusters can be reduced significantly (with $c$ chosen as low as $2^{-3}$) without loss in performance, unless many non-coreferring pairs are used (high $k$). This indicates that, for \ac{ecbp}, there is little benefit in generating all possible coreferring mention pairs for training, and that achieving a broad selection of mention pairs from many different events is more important.
Based on these results, and taking into account the distribution of cluster sizes in each corpus (see \Cref{fig:cluster_size_distribution}), we chose $(c=8, k=8)$ for \ac{ecbp} and the similarly distributed \ac{gvc} in the main experiments. For experiments involving \ac{fcct}, we chose $(c=2, k=8)$ to reduce the impact of its few large, mostly redundant clusters on training.

\printacronyms[include=main]
\printacronyms[name={Corpus Acronyms}, include=corpus, display=all]
\printacronyms[name={System Acronyms}, include=system, display=all]

\bibliographystyle{bibstyle}
\starttwocolumn
\bibliography{bib.bib}

\end{document}